\documentclass[11pt]{article}
\usepackage{fullpage}

\usepackage{amsmath,amsfonts,amssymb}
\usepackage{algorithm,algorithmic}
\usepackage{multirow}
\usepackage{cellspace}
\usepackage{setspace}
\usepackage{bm}
\usepackage{bbm}
\usepackage{subfigure}
\usepackage{graphicx}
\usepackage{tikz,pgfplots}
\usepackage{framed}
\usepackage{amsthm}

\usepackage[round]{natbib}
\setcitestyle{authoryear,round,citesep={;},aysep={,},yysep={;}}

\usepackage{hyperref}
\hypersetup{colorlinks,
            linkcolor=blue,
            citecolor=blue,
            urlcolor=magenta,
            linktocpage,
            plainpages=false}

\usepackage[capitalise]{cleveref}

\usepackage{times}
\usepackage{graphicx} 
\usepackage{subfigure} 

\usepackage{algorithm}
\usepackage{algorithmic}

\usepackage{hyperref}

\usepackage{geometry}                
\usepackage[parfill]{parskip}    
\usepackage{subfigure}
\usepackage{graphicx}
\usepackage{amssymb}
\usepackage{epstopdf}
\usepackage{algorithm}
\usepackage{algorithmic}
\usepackage[mathcal]{eucal}

\newtheorem{proposition}{Proposition}[section]
\newtheorem{theorem}{Theorem}[section]
\newtheorem{definition}{Definition}[section]
\newtheorem{lemma}{Lemma}[section]

\newcommand{\tF}{\tilde{F}}

\newcommand{\tg}{\tilde{g}}

\newcommand{\1}[1]{1_{ \{#1\}}}

\renewcommand{\le}{~\leq~}

\DeclareMathOperator*{\argmin}{arg\,min}
\DeclareMathOperator*{\argmax}{arg\,max}

\def\reals{{\mathcal R}}

\newcommand{\K}{\mathcal{K}}
\newcommand{\R}{\mathcal{R}}

\newcommand{\ignore}[1]{}

\def\reals{{\mathbb R}}

\def\V{{\mathcal V}}

\def\bold0{\mathbf{0}}

\newcommand\E{\mbox{\bf E}}

 \newcommand{\amax}{{a}}
 
 \newcommand{\Dr}{\mathcal{D}_{\mathcal{R}}}
  
  \newcommand{\dg}{{\rm DualGap}}

\renewcommand{\tg}{\tilde{g}}
\newcommand{\tM}{\tilde{M}}
\renewcommand{\tF}{\tilde{F}}
\newcommand{\ba}{a_0}

\def\1{\mathbf{1}}

\def\1{\mathbf{1}}

\newcommand{\regret}{\text{Regret}}

\newcommand{\rA}{\rm{(A)}}
\newcommand{\rB}{\rm{(B)}}
\newcommand{\rC}{\rm{(C)}}
\newcommand{\rD}{\rm{(D)}}
\newcommand{\rE}{\rm{(E)}}

\newcommand{\U}{\mathcal{U}}
\renewcommand{\V}{\mathcal{V}}

\newtheorem*{lemma*}{Lemma}
\newtheorem*{proposition*}{Proposition}
\newtheorem*{theorem*}{Theorem}

\title{A Universal Algorithm for Variational Inequalities Adaptive to Smoothness and Noise}%

\author{%
Francis Bach\footnote{INRIA, ENS, PSL Research University
Paris, France.~
Email:~\texttt{francis.bach@inria.fr}.} 
\and
Kfir Y. Levy\footnote{Department of Computer Science, ETH Z\"urich.~
Email:~\texttt{yehuda.levy@inf.ethz.ch}.}
}

\begin{document}
\maketitle

\begin{abstract}
We consider variational inequalities coming from monotone operators, a setting that includes convex minimization and convex-concave saddle-point problems. 
We assume an access to potentially noisy unbiased values of the monotone operators and assess convergence through a compatible gap function which corresponds to the standard optimality criteria in the aforementioned subcases.
We present a  universal algorithm for these inequalities based on the Mirror-Prox algorithm.
Concretely, our algorithm \emph{simultaneously} achieves the optimal rates for the smooth/non-smooth, and noisy/noiseless settings. This is done without any prior knowledge of these properties,
and in the  general set-up of arbitrary norms and compatible Bregman divergences. For convex minimization and convex-concave saddle-point problems, this leads to new adaptive algorithms.
Our method relies on a novel yet simple adaptive choice of the step-size, which can be seen as  the appropriate  extension of AdaGrad to  handle constrained problems.
\end{abstract}

\newpage
\section{Introduction}
Variational inequalities are a classical and general framework to encompass a wide variety of optimization problems such as convex minimization and convex-concave saddle-point problems, which are ubiquitous in machine learning and optimization \citep{nemirovski2004prox,juditsky2011solving,juditsky2016solving}.  Given a convex subset $\K$ of $\reals^d$, these inequalities are often defined from a \emph{monotone operator}  $F:\K\mapsto\reals^d$ (which we will assume single-valued for simplicity), such that for any $(x,y) \in \K \times \K$, $(x-y) \cdot ( F(x) - F(y) ) \geqslant 0$. The goal is then to find a strong solution $x^\ast \in \K$ to the variational inequality, that is, such that
\begin{equation}
\label{eq:vi}
\forall x \in \K,   \  (   x^\ast - x ) \cdot F(x^\ast)  \leq 0.
\end{equation}
For convex minimization problems, the operator $F$ is simply the subgradient operator, while for convex-concave saddle-point problems, the operator $F$ is composed of the subgradient with respect to the primal variable, and the negative supergradient with respect to the dual variables (see a detailed description in Section~\ref{Sec:GamesSetting}). In these two classical cases, solving the variational inequality corresponds to the usual notion of solution for these two problems. While our main motivation is to have a unique framework for these two subcases, the  variational inequality framework is more general (see e.g.~\citet{nemirovski2004prox} and references therein). 

In this paper we are interested in algorithms to solve the inequality in Eq.~\eqref{eq:vi}, while only accessing an oracle for $F(x)$ for any given $x \in \K$, or only an unbiased estimate of $F(x)$.
We also assume that we may efficiently project onto the set $\K$ (which we assume compact throughout this paper) using Bregman divergences.
In terms of complexity bounds, this problem is by now well-understood with matching upper and lower bounds in a variety of situations. In particular the notion of smoothness (i.e., Lipschitz-continuity of $F$ vs.~simply assuming that $F$ is bounded) and the presence of noise are the two important factors influencing the convergence rates. For example, the ``Mirror-Prox'' algorithm of~\citet{nemirovski2004prox} and \citet{juditsky2011solving}, given the correct step-size (that depends heavily on the properties of the problem, see Section~\ref{sec:SettingVariationalInseq}), attains the following bounds:
\begin{itemize}
\item For non-smooth problems where the operator (and its unbiased estimates) is  bounded by $G$, the rate $O( G D / \sqrt{T})$ is attained after $T$ iterations, where $D$ is the proper notion of diameter for the set $\K$. 
\item For smooth problems with $L$-Lipschitz operators, and a noise variance of $\sigma^2$, the convergence rate is $O( LD^2 / T +  {\sigma} D / \sqrt{T})$. 
\end{itemize}
These rates are actually optimal for this class of problems\footnote{The class of problems indeed includes convex optimization with lower bounds in $O(1/\sqrt{T})$~\citep{nemirovskii1983problem} and bilinear saddle-point problems with lower bound in $O(1/T)$~\citep{nemirovsky1992information}.}. However, practitioners may not know in which class their problem lies or know all the required constants needed for running the algorithms. Thus universal (sometimes called adaptive) algorithms are needed to leverage the potentially unknown properties of an optimization problem. Moreover, locally, the problem could be smoother or less noisy than globally, and thus classical algorithms would not benefit from extra local speed-ups.  

In this paper we make the following contributions:
\begin{itemize}
\item We present a universal algorithm for variational inequalities based on the Mirror-Prox algorithm,   for both deterministic and stochastic settings.
Our method employs a  simple adaptive choice of the step-size that leads to optimal rates for smooth and non-smooth variational inequalities. Our algorithm does not require prior knowledge regarding the smoothness or noise properties of the problem.
\item This is done in the  general set-up of arbitrary norms and compatible Bregman divergences.
\item For convex minimization and convex-concave saddle-point problems, this leads to new adaptive algorithms. 
 In particular, our new adaptive method can be seen as extension of AdaGrad~\citep{mcmahan2010adaptive,duchi2011adaptive}, that is more appropriate to handling constrained problems.
\end{itemize}
On the technical side, our work combines the Mirror-Prox method with a novel adaptive learning rate rule inspired by online learning techniques such as  AdaGrad~\citep{mcmahan2010adaptive,duchi2011adaptive}, and  optimistic OGD \citep{chiang2012online,rakhlin2013optimization}.

\paragraph{Related work.}
Algorithms for solving variational inequalities date back to  \citet{korpelevich1976extragradient} who was the first to suggest the \emph{extragradient} method.
The key idea behind this method is the following: in each round~$t$ we make two updates. First, we take a gradient step from the current iterate $y_t$, which leads to a point $y_{t+1/2}$. Then, instead of applying another gradient step starting in $y_{t+1/2}$, we go back to $y_t$ and take a step using the gradient of $y_{t+1/2}$, which leads to $y_{t+1}$.

The work of  \citet{korpelevich1976extragradient}  was followed by \cite{korpelevich1983extrapolational,noor2003new}, who further explored the asymptotic behaviour of such  {extragradient}-like algorithms. The seminal work of \citet{nemirovski2004prox} was the first to establish non-asymptotic convergence guarantees of such a method, establishing a rate of $O(LD^2/T)$ for smooth problems. Nemirovski's method named \emph{Mirror-Prox} was further explored by \citet{juditsky2011solving}, who analyze the stochastic setting,  and present a Mirror-Prox version that obtains a  rate of $O(LD^2 /T + \sigma D /\sqrt{T})$, where $\sigma^2$ is the variance of the noise terms. It is also known that in the non-smooth case,
Mirror-Prox  obtains a rate of $O(GD/\sqrt{T})$ \citep{juditsky2011second}.
Note that the Mirror-Prox versions that we have mentioned so far require prior knowledge about the smoothness/non-smoothness and on the noise properties of the problem (i.e., $\sigma$), in order to obtain 
the optimal bounds for each case\footnote{For the special case of bi-linear saddle-point problems, \citet{juditsky2013randomized} designed an algorithm that is adaptive to noise, but not to non-smoothness (which is irrelevant for bi-linear problems).}. Conversely, our method obtains these optimal rates without any such prior knowledge.  Note that  \cite{yurtsever2015universal,dvurechensky2018generalized} devise universal methods to solve variational inequalities that adapt to the smoothness of the problem. Nevertheless, these methods build on a line search technique that is inappropriate for handling noisy problems.  
Moreover, these methods require a predefined accuracy parameter   as an input, which requires careful hyperparameter tuning.

In the past years there have been several works on universal  methods for convex optimization (which is a particular case of the variational inequalities framework). 
\citet{nesterov2015universal} designed a universal method  that obtains the optimal convergence rates of $O(LD^2/T^2)$ and $O(GD/\sqrt{T})$ for smooth/non-smooth optimization, without any prior knowledge of the smoothness.
Yet, this method builds on a line search technique that is inappropriate to handling noisy problems.
Moreover, it also requires a predefined accuracy parameter as an input, which requires careful  tuning.

\citet{levy2017online}   designed alternative universal methods for convex minimization that do not require line search, yet these methods obtain a rate of $O(1/T)$ rather than the accelerated $O(1/T^2)$ rate for smooth objectives. Moreover, their results for the smooth case only holds for  \emph{unconstrained problems}. The same also applies  to the well known AdaGrad method  \citep{mcmahan2010adaptive,duchi2011adaptive}.
 Recently, \citet{levy2018online} have presented a universal method that obtains the optimal rates for smooth/non-smooth and noisy/noiseless settings, without any prior knowledge of these properties. Nevertheless, their results for the smooth case are only valid in the \emph{unconstrained setting}. Finally, note that these convex optimization methods are usually not directly applicable to the more general variational inequality framework.

Methods for solving convex-concave zero-sum games or saddle-point problems (another particular case of the variational inequality framework) were explored by the online learning community. The seminal work of 
\citet{freund1999adaptive} has shown how to employ regret minimization algorithms to solve such games at a rate of $O(1/\sqrt{T})$.
While the Mirror-Prox method solves such games at a faster rate of  $O(1/{T})$, it requires communication between the players. Interestingly, \citet{daskalakis2011near} have shown how to achieve a rate of $O(1/{T})$
without communication.
Finally, \citet{rakhlin2013optimization} have provided a much simpler  algorithm that obtains the same guarantees.

\section{Variational Inequalities and Gap Functions}
\label{sec:SettingVariationalInseq}
Here we present our general framework of variational inequalities with monotone operators, and introduce the notion of associated convex gap function. In Section~\ref{Sec:ConvexOptSetting} and \ref{Sec:GamesSetting}, we show how this framework captures the settings of convex optimization, as well as   convex-concave minimax games.
\paragraph{Preliminaries.}
 Let $\|\cdot\|$ be a general norm and $\|\cdot\|_*$ be its dual norm. 
A  function $f:\K\mapsto\reals$ is \emph{$\mu$-strongly convex} over a convex set $\K$, if for any $x\in\K$ and 
 any $\nabla f(x)$, a subgradient of $f$ at~$x$, 
\begin{align*}
&f(y) \geq f(x) + \nabla f(x) \cdot (y-x) + \frac{\mu}{2}\|x - y\|^2~;\quad \forall x,y \in \K.
 \end{align*}
A  function $f:\K\mapsto\reals$ is \emph{$L$-smooth} over  $\K$ if,~
$
\|\nabla f(x)-\nabla f(y)\|_* \leq L \|x-y\|~;\quad \forall x,y \in \K~.
$ 
Also, for a convex differentiable function $f(\cdot)$, we define its \emph{Bregman divergence} as follows,
$$
\mathcal{D}_f(x,y) = f(x) -f(y) - \nabla f(y)\cdot(x-y)~.
$$
Note that $\mathcal{D}_f(\cdot,\cdot)$ is always non-negative. For more properties, see, e.g.,~\citet{nemirovskii1983problem} and references therein.

\subsection{Gap functions}
We are considering a monotone operator $F$ from $\K$ to $\reals^d$, which is single-valued for simplicity\footnote{That is, each $x\in\K$ is mapped to a  \emph{single} $F(x) \in\reals^d$; we could easily extend to the multi-valued setting~\citep{bauschke2011convex}, at the expense of more cumbersome notations.}.
Formally,  a \emph{monotone operator} satisfies,
$$
(x-y) \cdot ( F(x) - F(y) ) \geq 0; \qquad
\forall (x,y) \in \K \times \K~.
$$
And we are usually looking for a \emph{strong} solution $x^\ast \in \K$ of the variational inequality, that satisfies
$$
\sup_{x \in \K} \ \  (   x^\ast - x ) \cdot F(x^\ast)  \leq 0.
$$
When $F$ is monotone, as discussed by~\citet{juditsky2016solving}, a strong solution is also a \emph{weak} solution, that is, $\sup_{x \in \K} \  (   x^\ast - x ) \cdot F(x)  \leq 0$. Note that we do not use directly the  monotonicity property of $F$; we only use the existence of a compatible gap function with respect to~$F$, 
which is an adapted notion of merit function to characterize convergence,
that we define in Def.~\ref{def:ProxOp}.
We show below that this definition captures the settings of  convex optimization and convex-concave games.

We thus assume that we are given a convex set $\K$, as well as  a \emph{gap function} $\Delta:\K\times\K\mapsto \reals$.  For a given solution  ${x}\in\K$, we define its \emph{duality gap} as follows,
\begin{align}\label{eq:DualityGapGeneral}
\dg(x): = \max_{y\in\K} \Delta({x},y)~.
\end{align}
We assume to  have an access to an oracle for $F$, i.e., upon querying this oracle with $x\in\K$, we receive $F(x)$. 
Our goal is to find a solution such that its duality gap is (approximately) zero. We  also consider a stochastic setting (similarly to \cite{juditsky2011solving}), where our goal  is to provide guarantees on the \emph{expected} duality gap. Next we present the \emph{central definition} of this paper:
 \begin{definition}[\textbf{Compatible gap function}]
\label{def:ProxOp}
Let $\K\subseteq \reals^d$ be a convex set, and let $\Delta: \K\times \K:\mapsto \reals$, such that $\Delta$ is convex with respect to its first argument.
We say that the function $\Delta$ is a gap function compatible with the monotone operator $F:\K\mapsto \reals^d$  if,
$$
\Delta(x,y) \leq F(x)\cdot(x-y), \qquad \forall x,y\in\K~,
$$
and $x^\ast \in \K$ is a solution of Eq.~\eqref{eq:vi} if and only if $\dg(x^\ast): = \max_{y\in\K} \Delta({x}^\ast,y)=0$.
\end{definition}
Note that given the notion of  solution to the variational inequality in Eq.~\eqref{eq:vi}, the function $(x,y) \mapsto F(x)\cdot(x-y)$ is a good candidate for $\Delta$, but it is not convex in $x$ in general and thus Jensen's inequality cannot be applied. 

\paragraph{Assumptions on $F$.}
Throughout this paper we will assume there exists a bound $G$ on the magnitude of $F$ (and all of its unbiased estimates), i.e.,
$$
\| F(x)\|_* \leq G, \qquad \forall x\in\K~.
$$
We will sometimes consider the extra assumption that   $F$ is $L$-smooth w.r.t.~a  given norm $\|\cdot\|$, i.e.,
$$
\| F(x)-F(y)\|_* \leq L \|x-y\|, \qquad \forall x,y\in\K~,
$$
where $\|\cdot\|_*$ is the dual norm of $\| \cdot\|$. Note that we define the notion of smoothness for functions $f:\K\mapsto \reals$, as well as to monotone operators $F:\K\mapsto \reals^d$ . These two different notions coincide when $F$ is the gradient of $f$ (see Sec.~\ref{Sec:ConvexOptSetting}).

Next we  show that the setting that we described in the section (see Def.~\ref{def:ProxOp}) captures two important settings, namely convex optimization and convex-concave zero-sum games.

\subsection{Convex Optimization}
\label{Sec:ConvexOptSetting}
Assume that $\K$ is a convex set, and $f:\K\mapsto \reals$ is convex over $\K$.
In the convex optimization setting our goal is to minimize $f$, i.e.,
$$
\min_{x\in\K} f(x)~.
$$
We assume that we may query   (sub)gradients of $f$.
Next we show how this setting is captured by the variational inequality setting.
Let us define a gap function and an operator $F$ as follows,
$$
\Delta(x,y): = f(x)-f(y), \quad \forall x,y,\in\K~, \qquad \& \qquad  F(x): = \nabla f(x), \quad \forall x\in\K.
$$
Then by the (sub)gradient inequality for convex functions, it immediately follows that $\Delta$ is a compatible gap function with respect to $F$. 
Also, it is clear that $\Delta(x,y)$ is convex with respect to $x$.
Finally, note that the duality gap in this case is the natural sub-optimality measure, i.e.,
$$
\dg(x) :=\max_{y\in\K}\Delta(x,y) =  f(x) - \min_{y\in\K}f(y)~.
$$
Moreover, if $f$ is $L$-smooth w.r.t. a norm $\| \cdot\|$, then  $F$ is smooth with respect to the same norm. 

\subsection{Convex-Concave Zero-sum  Games}
\label{Sec:GamesSetting}
 
Let  $\phi:\U \times \V \mapsto\reals$, where $\phi(u,v)$ is convex in $u$ and concave in $v$,  and $\U\subseteq \reals^{d_1},\V\subseteq \reals^{d_2},$ are compact convex sets.  
The convex-concave zero-sum game induced by $\phi$ is defined as follows, 
$$
\min_{u\in \U}\max_{v\in \V}\phi(u,v)~.
$$
The performance measure for such games is the duality gap which is defined as,
\begin{align}\label{eq:DG_games}
\dg(u,v) = \max_{v\in \V}\phi(u,v) - \min_{u\in \U}\phi(u,v)~.
\end{align} 
The duality gap is always non-negative, and we seek an (approximate) equilibrium, i.e., a point 
$(u^*,v^*)$ such that $\dg(u^*,v^*) =0$.

This setting can be classically described as a variational inequality problem.
Let us denote,
$$
x := (u,v) \in \U\times \V~; \quad  \mbox{ and } \quad \K: = \U\times \V~.
$$
For any $x=(u,v), x_0=(u_0,v_0)\in\K$, define a gap function and an operator $F:\K\mapsto \reals^{d_1+d_2}$, as follows,
$$
\Delta(x,x_0) := \phi(u,v_0) - \phi(u_0,v)~, \quad \mbox{ and } \quad F(x) : = (\nabla_u \phi(u,v), -\nabla_v \phi(u,v))~.
$$
It is immediate to show that this gap function, $\Delta$, induces the duality gap appearing in Eq.~\eqref{eq:DG_games}, i.e., $\dg(x) := \max_{x_0\in\K} \Delta(x,x_0)$.
Also, from the convex-concavity of $\phi$ it  immediately follows  that $\Delta(x,x_0)$ is convex in~$x$.
The next lemma from \citet{nemirovski2004prox} shows that  $\Delta$ is a gap function compatible with $F$ (for completeness we provide its proof in Appendix~\ref{sec:Proof_lem:GradIneqSaddleO}).

\vspace*{.25cm}

\begin{lemma}\label{lem:GradIneqSaddleO}
The following applies for any $x:=(u,v), x_0:=(u_0,v_0)\in \U\times \V$:
$$
\Delta(x,x_0):=\phi(u,v_0) - \phi(u_0,v) \leq  F(x) \cdot(x-x_0)~.
$$
\end{lemma} 
\paragraph{Mirror Map for Zero-sum Games.}
In this work, our variational inequality method employs a mirror-map over  $\K$.
For the case of zero-zum games $\K: =\U\times \V$, and we usually have separate mirror-map  terms,
$\R_\U:\U\mapsto \reals$, and $\R_\V:\V\mapsto \reals$. \citet{juditsky2011second}  have found a way  to appropriately define a mirror-map  over $\K$ using   $\R_\U,\R_\V$.  We hereby describe it.

Assume that the separate mirror-maps are $1$-strongly convex w.r.t.~norms 
$\| \cdot\|_\U$ and $\|\cdot\|_\V$, and let $\| \cdot\|_\U^*$ and $\|\cdot\|_\V^*$ be the respective dual norms.
Also, define $D_\U^2: = \max_{u\in\U}\R_\U(u) - \min_{u\in\U}\R_\U(u)$, and similarly define  $D_\V^2$.
 \citet{juditsky2011second}    suggest to employ, 
$$
\R_\K(x) = \frac{1}{D_\U^2}\R_\U(u) + \frac{1}{D_\V^2}\R_\V(v)~; \qquad \forall x: = (u,v)\in\K~,
$$
and to define,
\begin{align}\label{eq:NormK}
\|x\|_\K : = \sqrt{ \|u\|^2_\U/D_\U^{2} + \|v\|^2_\V/D_\V^{2} }~; \qquad \forall x: = (u,v)\in\K~.
\end{align}
In this case $\R_\K$ is $1$-strongly-convex w.r.t.~$\| \cdot\|_\K$. Also, the dual norm of $\|\cdot\|_\K$ in this case is,
\begin{align}\label{eq:DualNormK}
\|x\|_\K^* : = \sqrt{ D_\U^{2}(\|u\|_\U^*)^2 + D_\V^{2}(\|v\|_\V^*)^2 }~; \qquad \forall x: = (u,v)\in\reals^{d_1}\times \reals^{d_2}~.
\end{align}

\paragraph{Smooth Zero-sum Games.}
It can be  shown that if the gradient mapping $\nabla_u \phi(u,v)$, and $\nabla_v \phi(u,v)$  are Lipschitz-continuous with respect to both $u$ and $v$, then the monotone operator $F$ defined through
$F(x): = (\nabla_u \phi(u,v),-\nabla_v \phi(u,v))$ is also smooth.
Concretely, let $\|\cdot\|_\U$, and $\| \cdot\|_\V$ be norms over $\U$ and $\V$, and let 
$\|\cdot\|_\U^*$, and $\| \cdot\|_\V^*$ be their respective dual norms. \citet{juditsky2011second} show that  if the following holds $\forall u,u'\in\U, v,v'\in\V$, 
\begin{align*}
&\|\nabla_u \phi(u,v) -\nabla_u \phi(u',v)\|_\U^* \leq L_{11}\|u-u'\|_\U \\
&\|\nabla_u \phi(u,v) -\nabla_u \phi(u,v')\|_\V^* \leq L_{12}\|v-v'\|_\V \\
&\|\nabla_v \phi(u,v) -\nabla_v \phi(u,v')\|_\V^* \leq L_{22}\|v-v'\|_\V \\
&\|\nabla_v \phi(u,v) -\nabla_v \phi(u',v)\|_\U^* \leq L_{21}\|u-u'\|_\U~.
\end{align*}
Then it can be shown that $\forall x,x'\in\K$   
$$
\|F(x) - F(x')\|_\K^* \leq L \|x-x'\|_\K~,
$$
where $\|\cdot\|_K$, and $\|\cdot\|_\K^*$ are defined in Equations~\eqref{eq:NormK} and \eqref{eq:DualNormK}, and,
$$
L : =  2\max\{ L_{11}D_\U^2,L_{22}D_\V^2,L_{12}D_\U D_\V,L_{21}D_\U D_\V\}~.
$$


\section{Universal Mirror-Prox}
\label{sec:Offline}
This section presents our variational inequality algorithm. We first introduce the optimistic-OGD algorithm of \citet{rakhlin2013optimization}, and present its guarantees. Then we show how to adapt this algorithm together with a novel learning rate rule in order to solve variational inequalities in a universal manner. 
Concretely, we present an algorithm that, without any prior knowledge regarding the problem's smoothness, obtains a rate of $O(1/T)$  for smooth problems (Thm.~\ref{thm:Smooth}), and an  $O(\sqrt{\log T/T})$ rate  for non-smooth problems (Thm.~\ref{thm:NonSmooth}). 
Our algorithm can be seen as an adaptive version of the Mirror-Prox method \citep{nemirovski2004prox}. 

We provide a proof sketch of Thm.~\ref{thm:Smooth} in Section~\ref{sec:Sketchthm:Smooth}. The full proofs are deferred to the Appendix.

\subsection{Optimistic OGD}
Here we introduce the optimistic online gradient descent (OGD) algorithm of \citet{rakhlin2013optimization}.
This algorithm applies to the online linear optimization setting that can be described as a sequential game over $T$ rounds between a learner and an adversary.
In each round $t\in[T]$,
\begin{itemize}
\item the learner picks a decision point $x_t\in \K$,
\item the adversary picks a loss vector $g_t\in\reals^d$,
\item the learner incurs a loss of $g_t\cdot x_t$, and gets to view $g_t$ as a feedback.
\end{itemize}
The performance measure for the learner is the regret which is defined as follows,
$$
\regret: = \sum_{t=1}^Tg_t\cdot x_t - \min_{x\in \K}\sum_{t=1}^T g_t\cdot x~,
$$
and we are usually  interested in learning algorithms that ensure a regret which is sublinear in~$T$.

\paragraph{Hint Vectors.}
\citet{rakhlin2013optimization} assume that in addition to viewing the loss sequence  $\{g_t:\in \reals^d\}_{t}$, 
the learner may access a sequence of  ``hint vectors'' $\{M_t\in \reals^d\}_{t}$.
 Meaning that in each round $t$, prior to choosing $x_t$, the player gets to view a ``hint vector" $M_t \in\reals^d$.
 In the case where the hints are good predictions for the loss vectors, i.e., $M_t\approx g_t$, 
 \citet{rakhlin2013optimization} show that  this could be exploited to provide improved regret guarantees.
 Concretely, they suggest to use the following \emph{optimistic OGD} method:~
Choose
$y_0= \argmin_{x\in\K} \R(x)$, and $\forall t\geq 1$,
\begin{align}\label{eq:OptimisticGD}
x_t \gets \argmin_{x\in\K} M_t\cdot x + \frac{1}{\eta_t}\Dr(x,y_{t-1}),  \quad\mbox{ and } \quad
y_t \gets \argmin_{x\in\K} g_t \cdot x + \frac{1}{\eta_t}\Dr(x,y_{t-1}),
\end{align}
 where $\R(\cdot)$ is a $1$-strongly-convex function over $\K$ w.r.t.~a given norm $\|\cdot\|$,  and  $\Dr$ is the Bregman divergence of $\R$.
The following guarantees for  \emph{optimistic OGD} hold, assuming 
that the learning rate sequence is non-increasing (see proof in Appendix~\ref{sec:Proof_lem:RakhlinSridharan}) ,
\begin{lemma}[\cite{rakhlin2013optimization}]
\label{lem:RakhlinSridharan}
\begin{align}\label{eq:RegGurantees}
\!\!\! \regret  
&\leq
\frac{D^2}{\eta_1} + \frac{D^2}{\eta_T} + \sum_{t=1}^T\! \| g_t - M_t\|_* \cdot \|x_t-y_t\|
-\frac{1}{2}\sum_{t=1}^T \!\eta_t^{-1}\! \left(\|x_t-y_t\|^2 + \|x_t -y_{t-1}\|^2 \right),
\end{align}
where $D^2 = \max_{x \in \K} \R(x) - \min_{x \in \K} \R(x)$, and $\|\cdot\|_*$ is the dual norm of $\|\cdot\|$~.
\end{lemma}


 \subsection{Universal Mirror-Prox}
 Here we describe a new adaptive scheme for the learning rate of the above mentioned optimistic OGD. 
 Then we show that applying this adaptive scheme to solving variational inequalities yields an algorithm that adapts to smoothness and noise.  
\paragraph{A new Adaptive Scheme.}
 \cite{rakhlin2013optimization}  suggest to apply the following learning rate scheme inside optimistic OGD (Equation~\eqref{eq:OptimisticGD}),
$$
\eta_t = D/\max\Big\{\sqrt{\textstyle\sum_{t=1}^{t-1}\|g_t- M_t\|^2} +\sqrt{\textstyle \sum_{t=1}^{t-2}\|g_t- M_t\|^2} ,1 \Big\}.
$$ 
They show by employing this rule with a version of optimistic OGD yields an algorithm that  solves zero-sum matrix games at a fast  $O(1/T)$ rate, without any communication between the players.
While the Mirror-Prox algorithm \citep{nemirovski2004prox} achieves such a fast rate, it requires both players to communicate their iterates  to each other in every round.

Our goal here is different. We would like to adapt to the smoothness and noise of the objective, while allowing players to communicate.
To do so, we suggest to use the following adaptive scheme, 
\begin{align}\label{eq:LearningRate}
\eta_t =D/\sqrt{ G_0^2+\sum_{\tau=1}^{t-1}Z_\tau^2 },\qquad \text{where}~~ Z_\tau^2: = \frac{\|x_\tau-y_\tau\|^2 + \|x_\tau-y_{\tau-1}\|^2}{5\eta_\tau^2},
\end{align}
with the same definition of the diameter $D$ as in Lemma~\ref{lem:RakhlinSridharan}, and $G_0>0$ is an arbitrary constant.
Note that the best choice for $G_0$ is a tight upper bound on
   the dual norms of the $g_t$'s and $M_t$'s, which we denote here by $G$, i.e., $G : = \max_{t\in[T]}\max\{\|g_t\|_*,\|M_t\|_*\}$. Nevertheless, even if $G_0\neq G$ we still achieve convergence guarantees that scales with
   $$
   \alpha: = \max\{ G/G_0,G_0/G\}~.
   $$
 In this work we assume to know $D$, yet we do not assume any prior knowledge of $G$.

 Finally, note that $Z_\tau\in [0, G]$ for any $\tau\geq 1$; this immediately  follows by the next lemma. 
 \begin{lemma} \label{lemma:BoundZ}
 Let $G$ be a bound on the dual norms of  $\{g_t\}_t,\{M_t\}_t$. Then the above holds for 
 $y_{t-1},x_t,y_t$, that are used in Optimistic OGD (Eq.~\eqref{eq:OptimisticGD}), 
  \begin{align*}
 \|x_t-y_{t-1}\|/\eta_t\leq G, \quad \mbox{ and } \quad  \|y_t-y_{t-1}\|/\eta_t \leq G.
 \end{align*}
 \end{lemma}

\paragraph{Solving variational inequalities.}
So far we have described the online setting where the loss and hint vectors may change arbitrarily.  
Here we focus on the case where there exists a gap function $\Delta:\K\times\K\mapsto \reals$ that is compatible with a given monotone operator $F:\K\mapsto \reals^d$ (see Definition~\ref{def:ProxOp}). 
Recall that in this setting our goal is to minimize the duality gap induced by $\Delta$.
To do so, we choose  $g_t$ and $M_t$  in each round as follows,
\begin{align} \label{eq:MP_choices}
M_t = F(y_{t-1})~; \quad \mbox{ and } \quad g_t =F(x_{t})~,
\end{align}
These choices  correspond to the extragradient~\citep{korpelevich1976extragradient} and to  Mirror-Prox \citep{nemirovski2004prox} methods.

In Alg.~\ref{alg:UniMP} we present our universal Mirror-Prox algorithm for solving variational inequalities. 
This algorithm combines the Mirror-Prox algorithm (i.e., combining Eq.~\eqref{eq:MP_choices} inside the optimistic OGD of Eq.~\eqref{eq:OptimisticGD}), together with the new adaptive scheme that we propose in Eq.~\eqref{eq:LearningRate}.

\begin{algorithm}[t]
\caption{Universal Mirror-Prox  }
\label{alg:UniMP}
\begin{algorithmic}
\STATE \textbf{Input}: \#Iterations $T$, $y_0 =\argmin_{x\in\K} \R(x)$,  learning rate $\{\eta_t\} _{t}$ as in Eq.~\eqref{eq:LearningRate}
\FOR{$t=1 \ldots T$ }
\STATE {Set} $M_t = F(y_{t-1})$ 
\STATE {Update:}

\vspace*{-1cm}

\begin{align*}
x_t &\gets \argmin_{x\in\K} M_t\cdot x + \frac{1}{\eta_t}\Dr(x,y_{t-1}) , \quad \text{and define } \;g_t :=F(x_{t}), \\    
y_t &\gets \argmin_{x\in\K} g_t \cdot x + \frac{1}{\eta_t}\Dr(x,y_{t-1})
\end{align*}
\ENDFOR
\STATE {Output:}
$\bar{x}_T =  \frac{1}{T}\sum_{t=1}^{T}  x_{t}$
\end{algorithmic}
\end{algorithm}

\paragraph{Intuition.}
Before stating the guarantees of Alg.~\ref{alg:UniMP},    let us give some intuition behind   the learning rate that we suggest in Eq.~\eqref{eq:LearningRate}.
Note that the original Mirror-Prox algorithm employs two extreme learning rates for the non-smooth and smooth cases. In the smooth case the learning rate is \emph{constant}, i.e.,  $\eta_t \propto 1/L$,
 and in the non-smooth case it is decaying, i.e.,  $\eta_t \propto D/(G\sqrt{t})$. 
Next we show how our adaptive learning rate seems to implicitly adapts to the smoothness of the problem.

 For simplicity, let us focus on the convex optimization setting, where our goal is to minimize a convex function $f(\cdot)$, and therefore $F(x): = \nabla f(x)$. Also assume we use $\R(x): =\frac{1}{2} \|x\|_2^2$. 
 In this case,  optimistic OGD (Eq.~\eqref{eq:OptimisticGD}) is simply, $x_{t} \gets \Pi_\K(y_{t-1}-\eta_t M_t) $, and $y_{t} \gets \Pi_\K(y_{t-1}-\eta_t g_t)$, where $\Pi_\K$ is the orthogonal projection onto~$\K$. Now, let $x^* = \argmin_{x\in\K}f(x)$, and let us imagine two situations:\\
 \textbf{(i)} If $f(\cdot)$ is non-smooth around $x^*$, then the norms of the gradients are not decaying as we approach~$x^*$, and in this case the $\|Z_t\|$'s are lower bounded by some constant along all rounds.
This implies that $\eta_t $ will be proportional to $1/\sqrt{t}$.\\
\textbf{(ii)} Imagine that $f(\cdot)$ is smooth around $x^*$. If in addition $\nabla f(x^*) = 0$, this intuitively implies that the magnitudes $\|g_t\|$ and $\|M_t\|$ go to zero as we approach $x^*$, and therefore  $\|Z_t\|$'s will also go to zero. This intuitively means that $ \eta_t$ tends to a constant when $t$ tends to infinity. However, note that this behaviour can also be achieved by using an AdaGrad-like \citep{duchi2011adaptive} learning rate rule, i.e., $\eta_t \propto (\sum_{\tau=1}^t \|g_\tau\|^2 + \|M_\tau\|^2)^{-1/2}$.
The reason that we employ the more complicated learning rate of Eq.~\eqref{eq:LearningRate} is in order to handle the case where $f(\cdot)$ is smooth yet  $\|\nabla f(x^*)\| > 0$. In this case, the norms of $\|g_t\|$ and $\|M_t\|$ will not decay as we approach $x^*$; nevertheless the norms of $\|Z_t\|$'s will  intuitively go to zero, implying $\eta_t$ tends to a constant. Thus, in a sense, our new learning rate rule can be seen as the appropriate adaptation of AdaGrad to the constrained case.

\paragraph{Guarantees.}  We are now ready to state our guarantees.
 We show that  when the monotone operator $F$ is smooth, then we minimize the \emph{duality gap} in Eq.~\eqref{eq:DualityGapGeneral} at a fast rate of $O(1/T)$. Conversely, when the monotone operator is non-smooth, then we obtain a rate of $O(\sqrt{\log T/T})$.
This is achieved without any prior knowledge regarding the smoothness of $F$. 
 The next result addresses the smooth case (we provide a proof sketch in Sec.~\ref{sec:Sketchthm:Smooth}; the full proof appears in App.~\ref{sec:ProofTheoremSmooth}),
 \begin{theorem}\label{thm:Smooth}
 Assume that $F$ is  $L$-smooth, and $G$-bounded. Then Alg.~\ref{alg:UniMP} used with the learning rate of Eq.~\eqref{eq:LearningRate} implies the following bound,
$$
\dg(\bar{x}_T): = \max_{x\in\K}\Delta(\bar{x}_T,x) \leq O\Big( \frac{\alpha GD+\alpha^2 L  D^2+L  D^2\log(L  D/G_0)}{T}\Big)~.
$$
 \end{theorem}
 Recall that  $\alpha: = \max\{G/G_0,G_0/G \}$ measures the quality of our prior knowledge $G_0$ regarding the actual bound $G$ on the  norms of $F(\cdot)$.
 Next we present our guarantees  for the non-smooth case.
 \begin{theorem}\label{thm:NonSmooth}
 Assume that $F$ is  $G$-bounded.
 Alg.~\ref{alg:UniMP} used with the learning rate of Eq.~\eqref{eq:LearningRate} implies,
$$
\dg(\bar{x}_T): =\max_{x\in\K}\Delta(\bar{x}_T,x) \leq O\Big( {\alpha GD\sqrt{\log T}}/{\sqrt{T}}\Big)~.
$$
 \end{theorem}
Up to logarithmic terms, we recover the results from \cite{juditsky2011solving}, with a potential extra factor $\alpha$, which is equal to $1$ if we know a bound $G$ on the norms of the values of $F$ (but we do not require this value to obtain the correct dependence in $T$). The proof of Thm.~\ref{thm:NonSmooth} appears in App.~\ref{sec:Proofthm:NonSmooth}.

\subsection{Proof Sketch of Theorem~\ref{thm:Smooth}}
\label{sec:Sketchthm:Smooth}
\begin{proof}
 We shall require the following simple identity (see \cite{rakhlin2013optimization}),
 \begin{align*}
  \|g_t-M_t\|_* \cdot \|x_t-y_t\| = \min_{\rho>0}\Big\{ \frac{\rho}{2}\|g_t-M_t\|_*^2 + \frac{1}{2\rho}\|x_t-y_t\|^2\Big\}.
 \end{align*}
 Using the above with $\rho=1/L $, together with $g_t: = F(x_t), M_t = F(y_{t-1})$, and using the $L$-smoothness of $F$ gives,
 \begin{align*}
  \|g_t-M_t\|_* \cdot \|x_t-y_t\|
  &\leq
   \frac{L }{2}\|x_t-y_{t-1}\|^2 + \frac{L }{2}\|x_t-y_t\|^2 ~.
 \end{align*}
Combining this inside the regret bound of Eq.~\eqref{eq:RegGurantees}  and re-arranging we obtain,
 \begin{align} \label{eq:SmoothPart1Sketch}
 \regret 
&\leq
 \frac{D^2}{\eta_1} + \frac{D^2}{\eta_T} 
+\frac{5}{2}\sum_{t=1}^T\left(L  -\frac{1}{\eta_t}\right)\eta_t^2 Z_t^2~,
 \end{align}
 where we have used, $Z_t^2: = \left(\|x_t-y_t\|^2 + \|x_t-y_{t-1}\|^2\right)/{5\eta_t^2}$.\\
 Let us define $\tau_*: = \max\{t\in[T]:~ {1}/{\eta_t} \leq 2L \}$, and divide the last term of the regret as follows,
 \begin{align*}
\! \sum_{t=1}^T\Big(L  -\frac{1}{\eta_t}\Big)\eta_t^2 Z_t^2 
 &~=~
 \sum_{t=1}^{\tau_*} \Big(L  -\frac{1}{\eta_t}\Big)\eta_t^2 Z_t^2 
 +
  \sum_{\tau_*+1}^T \Big(L  -\frac{1}{\eta_t}\Big)\eta_t^2 Z_t^2  \\
 &~\leq~
 \sum_{t=1}^{\tau_*}L  \eta_t^2 Z_t^2 
 - 
  \frac{1}{2}\sum_{t=\tau_*+1}^T\eta_t Z_t^2 ~,
 \end{align*}
 where in the second line we use $2L \leq \frac{1}{\eta_t}$ which holds for $t> \tau_\star$; implying that $L -\frac{1}{\eta_t}\leq -\frac{1}{2\eta_t}$. 
 Plugging the above back into Eq.~\eqref{eq:SmoothPart1Sketch} we obtain,
  \begin{align} \label{eq:SmoothPart11Sketch}
 \regret 
&\leq
 \frac{D^2}{\eta_1} + 
  \underset{\rA}{\underbrace{\frac{D^2}{\eta_T} - \frac{5}{4}\sum_{t=\tau_*+1}^T\eta_t Z_t^2 }}
  +
  \underset{\rB}{\underbrace{\frac{5}{2}\sum_{t=1}^{\tau_*}L  \eta_t^2 Z_t^2  }}~.
 \end{align}
Next we bound terms $\rA$ and $\rB$ above.
To bound $\rA$ we will require the following lemma,
\begin{lemma*}\label{lem:SqrtSumReversedGeneralizedSketch}
For any non-negative numbers $a_1,\ldots, a_n\in [0,\amax]$, and $\ba\geq0$, the following holds:
\begin{equation*}
\sqrt{\ba+\sum_{i=1}^{n-1} a_i} -\sqrt{\ba}\leq \sum_{i=1}^n \frac{a_{i}}{\sqrt{\ba+\sum_{j=1}^{i-1} a_j}} \leq 
 \frac{2\amax}{\sqrt{\ba}}+3\sqrt{\amax} +3\sqrt{\ba +\sum_{i=1}^{n-1} a_i}~.
\end{equation*}
\end{lemma*}
Recalling that $\eta_t =D/\sqrt{ G_0^2+\sum_{\tau=1}^{t-1}Z_\tau^2 }$  ~~(see Eq.~\eqref{eq:LearningRate}), and also recalling that $Z_\tau\in [0,G]$ we can use the above lemma to bound term $\rA$,
 \begin{align} \label{eq:TermASketch}
 \!
 \rA&: 
 ~=~
 D\sqrt{G_0^2+\sum_{t=1}^{T-1} Z_t^2}- \frac{5D}{4}\sum_{t=\tau_*+1}^T\frac{Z_t^2}{\sqrt{G_0^2+\sum_{\tau=1}^{t-1}Z_\tau^2}} \nonumber\\
 &~\leq~
 DG_0+ D \sum_{t=1}^{T}\frac{Z_t^2}{\sqrt{G_0^2+\sum_{\tau=1}^{t-1}Z_\tau^2}}- \frac{5D}{4}\sum_{t=\tau_*+1}^T\frac{Z_t^2}{\sqrt{G_0^2+\sum_{\tau=1}^{t-1}Z_\tau^2}} \nonumber\\
 &~\leq~
DG_0+ D \sum_{t=1}^{\tau_*}\frac{Z_t^2}{\sqrt{G_0^2+\sum_{\tau=1}^{t-1}Z_\tau^2}} \nonumber\\
 &~\leq~
 3D(G+G_0) + {2DG^2}/{G_0} +3D^2\frac{1}{\eta_{\tau_*}}  \nonumber\\
 &~\leq~
 3D(G+G_0) + {2DG^2}/{G_0}+6L  D^2~,
 \end{align}
 where we have used the definition of $\tau_*$ which implies $1/\eta_{\tau*}\leq 2L $.
 
 \paragraph{Bounding term $\rB$:}
 In the full proof (Appendix~\ref{sec:ProofTheoremSmooth}) we show that ${\rB} \leq {O}(L D^2\log(L  D/G_0))$.  
  
\paragraph{Conclusion:} Combining the bounds on $\rA$ and $\rB$ into 
Eq.~\eqref{eq:SmoothPart11Sketch} and using $\eta_1=D/G_0$ implies, 
$$
\regret \leq O\left( \alpha DG+\alpha^2 L  D^2+L  D^2\log(L  D/G_0)\right)~,
$$
where we used the definition $\alpha: = \max\left\{{G}/{G_0},{G_0}/{G}\right\}$.
Combining the above with the definition of $\bar{x}_T$ and using Jensen's inequality (recall $\Delta$ is convex in its first argument), as well as with the fact that $\Delta$ is a compatible gap function w.r.t.~$F$ concludes the proof.
 \end{proof}

\section{Stochastic Setting}
\label{sec:Stochastic}

In this section we present the stochastic variational inequality setting. 
Then we show that using the exact same universal Mirror-Prox algorithm (Alg.~\ref{alg:UniMP}) that we have presented in the previous section, enables to provide the optimal guarantees for the stochastic setting. This is done without any prior knowledge regarding the  smoothness or the stochasticity of the problem.

\paragraph{Setting.}
The stochastic setting is similar to the deterministic setting that we have described in Sec.~\ref{sec:SettingVariationalInseq}. The only difference is that we do not have an access to the exact values of $F$.
Instead, we assume that when querying a point $x\in\K$ we receive an unbiased noisy estimate of the exact monotone mapping $F(x)$. More formally,  we have an access to an oracle   $\tilde{F}:\K\mapsto \reals^d$, such that for any $x\in\K$ we have,
 $$
 \E[\tilde{F}(x)\vert x] = F(x)~.
 $$
 We also assume to have a bound $G$ on the dual norms of $\tF$, i.e., almost surely,
 $
 \|\tilde{F}(x)\|_{*} \leq G; \; \forall x\in\K~.
 $
We are now ready to state our guarantees. Up to logarithmic terms, we recover the results from \cite{juditsky2011solving} with an universal algorithm that does not need the knowledge of the various regularity constants.
The first results regards the non-smooth noisy case.
\begin{theorem}\label{thm:NonSmoothNoisy}
 Assume that we receive  unbiased (noisy)  estimates $\tF$  instead of $F$ inside Alg.~\ref{alg:UniMP}. 
  Then Alg.~\ref{alg:UniMP} used with the learning rate of Eq.~\eqref{eq:LearningRate} ensures the following,
$$
\E\left[\dg(\bar{x}_T)\right]: =\E\max_{x\in\K}\Delta(\bar{x}_T,x) \leq O\Big( {\alpha GD\sqrt{\log T}}/{\sqrt{T}}\Big)~.
$$
 \end{theorem}
Next we further assume a bound on the variance of $\tF$, i.e., 
$
\E\left[\|\tilde{F}(x) - F(x)\|_*^2 \vert x\right] \leq \sigma^2, \; \forall x\in\K,
$
but we do not assume any prior knowledge of $\sigma$.
 The next theorem regards the smooth noisy case.
 \begin{theorem}\label{thm:SmoothNoisy}
 Assume that $F$ is  $L $-smooth, and assume  that we receive  unbiased (noisy)  estimates $\tF$  instead of $F$ inside Alg.~\ref{alg:UniMP}.  Then Alg.~\ref{alg:UniMP} used with the learning rate of Eq.~\eqref{eq:LearningRate} ensures the following,
\begin{align*}
\E\left[\dg(\bar{x}_T)\right]: &=\E\max_{x\in\K}\Delta(\bar{x}_T,x) \\
&\leq
 O\Big(\frac{\alpha GD+\alpha^2 L  D^2+L  D^2\log(L  D/G_0)}{T} 
+ \frac{\alpha \sigma D\sqrt{\log T}}{\sqrt{T}}\Big)~.
\end{align*}
 \end{theorem}
\subsection{Proof Sketch of Theorem~\ref{thm:NonSmoothNoisy}}
\begin{proof}
Let us denote by $\tg_t$ the noisy estimates of $g_t: = F(x_t)$.
Following the exact steps as in the proof of Theorem~\ref{thm:NonSmooth} implies the following holds  w.p.~1,
 $$
 \sum_{t=1}^T\tg_t\cdot(x_t-x)  \leq O(\alpha GD\sqrt{T\log T})~.
 $$
Recalling the definition of $\bar{x}_T$, and using Jensen's inequality implies that for any $x\in\K$,
 \begin{align} \label{eq:NoisyDeltaSketch}
T\cdot \Delta(\bar{x}_T,x)
& ~\leq~
\sum_{t=1}^T \Delta({x}_t,x) 
~\leq~
 \sum_{t=1}^Tg_t\cdot(x_t-x) \nonumber\\
 & ~=~
  \sum_{t=1}^T\tg_t\cdot(x_t-x) -  \sum_{t=1}^T\zeta_t\cdot(x_t-x) \nonumber\\
  &~\leq~
O(\alpha  GD\sqrt{T\log T}) -  \sum_{t=1}^T\zeta_t\cdot(x_t-x)~,
 \end{align}
 where we denote,
 $
 \zeta_t: = \tg_t-g_t .
 $
 And clearly $\{ \zeta_t\}_t$ is a martingale difference sequence.
Let $x^*: = \argmax_{x\in\K}\Delta(\bar{x}_t,x)$.   Taking $x = x^*$ and taking expectation over Eq.~\eqref{eq:NoisyDeltaSketch} gives,
\begin{align*}
T\cdot \E\Delta(\bar{x}_t,x^*)
&~\leq~ 
O(\alpha GD\sqrt{T\log T}) -  \E\sum_{t=1}^T\zeta_t\cdot(x_t-x^*)  \\
&~=~
O(\alpha GD\sqrt{T\log T})+  \E\sum_{t=1}^T\zeta_t\cdot x^*~. 
 \end{align*}
To establish the proof we are left to show that $\E\sum_{t=1}^T \zeta_t \cdot x^*\leq O(GD\sqrt{T})$.
This is challenging since $x^*$, by its definition, is a random variable that may depend on $\{\zeta_t\}_t$, implying that $\zeta_t \cdot x^*$ is not zero-mean. Nevertheless, we are able to make use of the 
 martingale difference property of  $\{\zeta_t\}_t$ in order to bound $\E\sum_{t=1}^T \zeta_t \cdot x^*$.
 This is done using the following proposition, 
\begin{proposition*}
Let $\K \subseteq \reals^d$ be a convex set, and $\R:\K\mapsto \reals$ be a $1$-strongly-convex function w.r.t. a norm $\|\cdot\|$ over $\K$. Also assume that $\forall x\in\K;\;\R(x) - \min_{x\in\K} \R(x)\leq \frac{1}{2}D^2$.
Then for any martingale difference sequence $(Z_i)_{i=1}^n\in \reals^d$, and any random vector $X$ defined over $\K$, we have,
$$ \E \Big[ \Big(  \sum_{i=1}^n Z_i \Big)^\top X  \Big]
\leq
 \frac{D  } { 2  }\sqrt{ \sum_{i=1}^n \E \|Z_i\|_*^2}.
$$
\end{proposition*}
We stress that the proposition  applies for random vectors $X$ which might even depend on $(Z_i)_{i=1}^n$.
\end{proof}

 \section{Conclusion}
In this paper, we have presented a universal algorithm for variational inequalities, that can adapt to smoothness and noise, leading, with a single algorithm with very little knowledge of the problem  to the best convergence rates in all these set-ups (up to logarithmic factors). There are several avenues worth exploring:  {(a)} an extension to a Matrix-AdaGrad-like algorithm~\citep{mcmahan2010adaptive,duchi2011adaptive} where a matrix gain is employed rather than a scalar step-size,  {(b)} an extension that could handle composite problems through additional proximal operators,  {(c)}  extensions of adaptivity to all H\"older-continuous mappings~\citep{dvurechensky2018generalized}, and finally  {(d)}  the inclusion of deterministic error terms to allow biased operator evaluations.

\subsection*{Acknowledgement}
We  would like to thank Nicolas Flammarion for fruitful discussions related to this work.

We acknowledge support from the European Research Council (grant SEQUOIA 724063), as well as from  the ETH Z\"urich Postdoctoral Fellowship and Marie Curie Actions for People COFUND program.

\bibliography{bib}
\bibliographystyle{apalike}

\newpage
\appendix
\paragraph{Appendix Description:}
In Appendix~\ref{sec:AppDeterministic}, we provide the missing proofs 
related to the Deterministic Setting (Section~\ref{sec:Offline}): the proofs of  Thm.~\ref{thm:Smooth} and Thm.~\ref{thm:NonSmooth} appear in App.~\ref{sec:ProofTheoremSmooth} and  \ref{sec:Proofthm:NonSmooth}. We also provide the proofs of Lemma~\ref{lem:GradIneqSaddleO} (see App~\ref{sec:Proof_lem:GradIneqSaddleO}), and Lemma~Lemma~\ref{lemma:BoundZ} (see App.~\ref{sec:ProoflemmaBoundZ}).

In Appendix~\ref{sec:AppStochastic}, we provide the missing proofs 
related to the Stochastic Setting (Section~\ref{sec:Stochastic}): the proofs of Thm.~\ref{thm:NonSmoothNoisy} and Thm.~\ref{thm:SmoothNoisy} appear in App.~\ref{sec:Proof_thm_NonSmoothNoisy} and \ref{Proof_thm:SmoothNoisy}. And we also prove  Proposition~\ref{prop:Concentration} (see App.~\ref{sec:Proof_prop:Concentration}), which is a central tool in the proofs of the stochastic case.
In Appendix~\ref{sec:AppC} we provide the remaining proofs for the paper.

 \section{Proofs for the Deterministic Setting (Section~\ref{sec:Offline})}
 \label{sec:AppDeterministic}
\subsection{Proof of Lemma~\ref{lem:GradIneqSaddleO}}
\label{sec:Proof_lem:GradIneqSaddleO}
\begin{proof} 
Using convexity we get for any $u_0\in \U, v\in\V$,
$$
\phi(u,v) - \phi(u_0,v) \leq \nabla_u \phi(u,v)\cdot (u-u_0)~.
$$
Similarly, using  concavity we  get for any $u\in\U, v_0\in \V$,
$$
-\phi(u,v) + \phi(u,v_0) \leq -\nabla_v \phi(u,v)\cdot (v-v_0)~.
$$
Summing both of the above equations gives for any $x:= (u,v),x_0: = (u_0,v_0)$,
$$
 \phi(u,v_0)- \phi(u_0,v) \leq F(x) \cdot (x-x_0)~.
$$
where we used $F(x): = (\nabla_u \phi(u,v),-\nabla_v \phi(u,v))$. This concludes the proof.
\end{proof}

 \subsection{Proof of Lemma~\ref{lemma:BoundZ}}
 \label{sec:ProoflemmaBoundZ}
 \begin{proof}
 Here we show that $\|x_t-y_{t-1}\|/\eta_t\leq G$; the proof of $\|y_t-y_{t-1}\|/\eta_t \leq G$ follows the exact same steps.
 
Before we start, note that the following holds for Bregman Divergences,
 $$
 \nabla_x \Dr(x,y) = \nabla \R(x)- \nabla \R(y)~.
 $$
 Now, the following applies for any $x\in\K$ by the definition of $x_t$ in Eq.~\eqref{eq:OptimisticGD},
 $$
 \nabla_{x:=x_t}(\eta_t M_t\cdot x + \Dr(x,y_{t-1}))\cdot(x-x_t) \geq 0~.
 $$
 Taking $x=y_{t-1}$, the above implies,
 $$
 \left( \eta_t M_t + \nabla\R(x_t) - \nabla\R(y_{t-1})\right) \cdot (y_{t-1}-x_t)\geq 0~.
 $$
 Combing this with $\R$ being $1$-strongly-convex w.r.t. $\|\cdot\|$ we obtain,
 $$
\eta_t M_t \cdot (y_{t-1}-x_t) \geq \left( \nabla\R(y_{t-1}) -  \nabla\R(x_{t}) \right) \cdot (y_{t-1}-x_t)\geq \|x_t - y_{t-1}\|^2~.
 $$
 Using  Cauchy-Swartz immediately implies that,
 $$
 \|x_t - y_{t-1}\|^2 \leq \eta_t\|x_t - y_{t-1}\|\cdot \|M_t\|_* \leq \eta_t\|x_t - y_{t-1}\|\cdot G~.
 $$
 Dividing the above equation by $\|x_t - y_{t-1}\|$ concludes the proof.
 \end{proof}

\subsection{Proof of Theorem~\ref{thm:Smooth}}
\label{sec:ProofTheoremSmooth}
 \begin{proof}
 We shall require the following simple identity (see \cite{rakhlin2013optimization}),
 \begin{align*}
  \|g_t-M_t\|_* \cdot \|x_t-y_t\| = \min_{\rho>0}\left\{ \frac{\rho}{2}\|g_t-M_t\|_*^2 + \frac{1}{2\rho}\|x_t-y_t\|^2\right\}.
 \end{align*}
 Using the above with $\rho=1/L $ we get,
 \begin{align}\label{Eq:rel111}
  \|g_t-M_t\|_* \cdot \|x_t-y_t\|
   &\leq \frac{1}{2L }\|g_t-M_t\|_*^2 + \frac{L }{2}\|x_t-y_t\|^2  \nonumber\\
  &=
 \frac{1}{2L }\|F(x_t)- F(y_{t-1})\|_*^2 + \frac{L }{2}\|x_t-y_t\|^2  \nonumber\\
  &\leq
   \frac{L }{2}\|x_t-y_{t-1}\|^2 + \frac{L }{2}\|x_t-y_t\|^2~,
 \end{align}
 where the last line uses the $L $-smooth of the operator $F$, i.e., $\|F(x_t)- F(y_{t-1})\|_* \leq L  \|x_t-y_{t-1}\|$. 
 Combining the above inside the regret bound of Eq.~\eqref{eq:RegGurantees} we obtain,
 \begin{align} \label{eq:SmoothPart1A}
 \regret 
 &\leq
 \frac{D^2}{\eta_1} + \frac{D^2}{\eta_T} 
+\frac{1}{2}\sum_{t=1}^T\left(L  -\frac{1}{\eta_t}\right)\left(\|x_t-y_t\|^2 + \|x_t -y_{t-1}\|^2 \right) \nonumber\\
&=
 \frac{D^2}{\eta_1} + \frac{D^2}{\eta_T} 
+\frac{5}{2}\sum_{t=1}^T\left(L  -\frac{1}{\eta_t}\right)\eta_t^2 Z_t^2~,
 \end{align}
 where we have used, $Z_t^2: = \left(\|x_t-y_t\|^2 + \|x_t-y_{t-1}\|^2\right)/{5\eta_t^2}$.\\

 Now let us define $\tau_*: = \max\{t\in[T]:~ {1}/{\eta_t} \leq 2L \}$. We can now divide the last term of the regret according to $\tau_*$,
 \begin{align*}
 \sum_{t=1}^T\left(L  -\frac{1}{\eta_t}\right)\eta_t^2 Z_t^2 
 &=
 \sum_{t=1}^{\tau_*} \left(L  -\frac{1}{\eta_t}\right)\eta_t^2 Z_t^2 
 +
  \sum_{\tau_*+1}^T \left(L  -\frac{1}{\eta_t}\right)\eta_t^2 Z_t^2  \\
 &\leq
 \sum_{t=1}^{\tau_*}L  \eta_t^2 Z_t^2 
 - 
  \frac{1}{2}\sum_{t=\tau_*+1}^T\eta_t Z_t^2
 ~.
 \end{align*}
 where in the second line we use $2L \leq \frac{1}{\eta_t}$ which holds for $t> \tau_\star$, implying that $L -\frac{1}{\eta_t}\leq -\frac{1}{2\eta_t}$. 
 Plugging the above back into Eq.~\eqref{eq:SmoothPart1A} we obtain,
 
  \begin{align} \label{eq:SmoothPart1}
 \regret 
&\leq
 \frac{D^2}{\eta_1} + 
  \underset{\rA}{\underbrace{\frac{D^2}{\eta_T} - \frac{5}{4}\sum_{t=\tau_*+1}^T\eta_t Z_t^2 }}
  +
  \underset{\rB}{\underbrace{\frac{5}{2}\sum_{t=1}^{\tau_*}L  \eta_t^2 Z_t^2  }}~.
 \end{align}
Next we bound terms $\rA$ and $\rB$ above,

\paragraph{Bounding term $\rA$:}

We will require the following lemma which we prove in Appendix~\ref{sec:Proof_SqrtSumReversedGeneralized},
\begin{lemma}\label{lem:SqrtSumReversedGeneralized}
For any non-negative numbers $a_1,\ldots, a_n\in [0,\amax]$, and $\ba\geq0$, the following holds:
\begin{equation*}
\sqrt{\ba+\sum_{i=1}^{n-1} a_i} -\sqrt{\ba}\leq \sum_{i=1}^n \frac{a_{i}}{\sqrt{\ba+\sum_{j=1}^{i-1} a_j}} \leq 
 \frac{2\amax}{\sqrt{\ba}}+3\sqrt{\amax} +3\sqrt{\ba +\sum_{i=1}^{n-1} a_i}~.
\end{equation*}
\end{lemma}

Recalling the learning rate rule that we use, $\eta_t =D/\sqrt{ G_0^2+\sum_{\tau=1}^{t-1}Z_\tau^2 }$  ~~(see Eq.~\eqref{eq:LearningRate}), and also recalling that $Z_\tau\in [0,G]$ we can use the above lemma to bound term $\rA$,
 \begin{align} \label{eq:TermA}
 \rA&: = \frac{D^2}{\eta_T} - \frac{5}{4}\sum_{t=\tau_*+1}^T\eta_t Z_t^2 \nonumber\\
 &=
 D\sqrt{G_0^2+\sum_{t=1}^{T-1} Z_t^2}- \frac{5}{4}\sum_{\tau_*+1}^T\eta_t Z_t^2 \nonumber\\
 &\leq
 DG_0+ D \sum_{t=1}^{T}\frac{Z_t^2}{\sqrt{G_0^2+\sum_{\tau=1}^{t-1}Z_\tau^2}}- \frac{5}{4}\sum_{t=\tau_*+1}^T\eta_t Z_t^2 \nonumber\\
 &=
 DG_0+  \sum_{t=1}^{T}\eta_t {Z_t^2}- \frac{5}{4}\sum_{t=\tau_*+1}^T\eta_t Z_t^2 \nonumber\\
 &\leq
 DG_0+ \sum_{t=1}^{\tau_*}\eta_t {Z_t^2}\nonumber\\
 &=
DG_0+ D \sum_{t=1}^{\tau_*}\frac{Z_t^2}{\sqrt{G_0^2+\sum_{\tau=1}^{t-1}Z_\tau^2}} \nonumber\\
 &\leq
DG_0+ \frac{2DG^2}{G_0}+3DG+3D\sqrt{G_0^2+\sum_{t=1}^{\tau^*-1}Z_t^2}\nonumber\\
 &\leq
 3D(G+G_0) + \frac{2DG^2}{G_0} +3D^2\frac{1}{\eta_{\tau_*}}\nonumber\\
 &\leq
 3D(G+G_0) + \frac{2DG^2}{G_0}+6L  D^2~,
 \end{align}
 where we have used the definition of $\tau_*$ which implies $1/\eta_{\tau*}\leq 2L $.

 \paragraph{Bounding term $\rB$:}
 We will  require the following lemma
 (proof is found in Appendix~\ref{app:Prooflem:Log_sum}),
\begin{lemma} \label{lem:Log_sum}
For any non-negative real numbers $a_1,\ldots, a_n\in[0,\amax]$, and $\ba \geq 0$,
\begin{align*}
\sum_{i=1}^n \frac{a_i}{\ba+\sum_{j=1}^{i-1} a_j} 
\le 
2+\frac{4\amax}{\ba}+2\log\left( 1+\sum_{i=1}^{n-1} a_i/\ba\right) ~.
\end{align*}
\end{lemma}

Recalling the learning rate rule that we use, $\eta_t =D/\sqrt{ G_0^2+\sum_{\tau=1}^{t-1}Z_\tau^2 }$  ~~(see Eq.~\eqref{eq:LearningRate}), and also recalling that $Z_\tau\in [0,G]$ we can use the above lemma to bound term $\rB$,
\begin{align}\label{eq:TermB}
\rB &= \frac{L }{2}\sum_{t=1}^{\tau_*} \eta_t^2 Z_t^2  \nonumber\\
&=
\frac{L  D^2}{2}\sum_{t=1}^{\tau_*}  \frac{Z_t^2 }{G_0^2+\sum_{\tau=1}^{t-1}Z_\tau^2 } \nonumber\\
&\leq
L  D^2 +2L  D^2 \frac{G^2}{G_0^2} +L  D^2 \log\left( \frac{G_0^2+\sum_{t=1}^{\tau_*-1}Z_t^2}{G_0^2}\right) \nonumber\\
&\leq
3L  D^2\max\{1,G^2/G_0^2\}+L  D^2 \log\left( \frac{(D/G_0)^2}{\eta_{\tau_*}^2}\right) \nonumber\\
&=
3L  D^2\max\{1,G^2/G_0^2\}+2L  D^2 \log\left( 2L  D/G_0\right)~,
\end{align}
where we have used the definition of $\tau_*$ which implies $1/\eta_{\tau*}\leq 2L $.

\paragraph{Conclusion:} Combining Equations \eqref{eq:TermA} and \eqref{eq:TermB} into 
Eq.~\eqref{eq:SmoothPart1} and using $\eta_1=D/G_0$ implies the following regret bound for Alg.~\ref{alg:UniMP},
$$
\regret \leq O\left( \alpha DG+\alpha^2 L  D^2+L  D^2\log(L  D/G_0)\right)~,
$$
where we used the definition $\alpha: = \max\left\{{G}/{G_0},{G_0}/{G}\right\}$.
Combining the above with the definition of $\bar{x}_T$ and using Jensen's inequality implies,
$$
\Delta(\bar{x}_T,x)  \leq O\left( \frac{\alpha DG+\alpha^2 L  D^2+L  D^2\log(L  D/G_0)}{T}\right), \qquad \forall x\in\K~.
$$  
 where we also used the fact that $\forall x_t,x\in\K$ the following holds,
 $$
 \Delta(x_t,x) \leq F(x)\cdot(x_t-x):= g_t\cdot(x_t-x)~.
 $$
 \end{proof}
 
 \newpage
  \subsection{Proof of Theorem~\ref{thm:NonSmooth}}
  \label{sec:Proofthm:NonSmooth}
  \begin{proof}
 Recall the regret bound of  Eq.~\eqref{eq:RegGurantees},
 
 \begin{align}\label{eq:RegGuranteesNonSmooth}
\regret&\leq
\underset{\rA}{\underbrace{\frac{D^2}{\eta_1} + \frac{D^2}{\eta_T} }}
-
 \underset{\rB}{\underbrace{\frac{1}{2}\sum_{t=1}^T \eta_t^{-1}\left(\|x_t-y_t\|^2 + \|x_t -y_{t-1}\|^2 \right)}}~
 + 
 \underset{\rC}{\underbrace{ \sum_{t=1}^T \| g_t - M_t\|_* \cdot \|x_t-y_t\| }}~.
\end{align}
Next we separately bound each of the above terms,
 \paragraph{ Bounding $\rA$:}
 Recalling the learning rate that we employ (see Eq.~\eqref{eq:LearningRate}) we have,
 \begin{align}\label{eq:TermA_Non}
 \rA 
 &:=
  \frac{D^2}{\eta_1} + \frac{D^2}{\eta_T} 
 =
 DG_0  + D\sqrt{ G_0^2+\sum_{t=1}^{T-1}Z_t^2 }~.
 \end{align}
 
\paragraph{ Bounding $\rB$:}
 Recalling Lemma~\ref{lem:SqrtSumReversedGeneralized} we may bound this term as follows,
  \begin{align}\label{eq:TermB_Non}
 \rB 
 &:=
 \frac{1}{2}\sum_{t=1}^T \eta_t^{-1}\left(\|x_t-y_t\|^2 + \|x_t -y_{t-1}\|^2 \right)  \nonumber\\
 &=
 \frac{5}{2}\sum_{t=1}^T \eta_t Z_t^2 \nonumber\\
 &=
 \frac{5D}{2}\sum_{t=1}^T \frac{Z_t^2}{\sqrt{ G_0^2+\sum_{\tau=1}^{t-1}Z_\tau^2 }} \nonumber\\
 &\geq
 \frac{5D}{2}\sqrt{ G_0^2+\sum_{t=1}^{T-1}Z_t^2 } - \frac{5DG_0}{2}~,
 \end{align}
where  the second line uses $Z_t^2: = \left(\|x_t-y_t\|^2 + \|x_t-y_{t-1}\|^2\right)/{5\eta_t^2}$.
 
 \paragraph{ Bounding $\rC$:}
Lets us recall  Lemma~\ref{lem:Log_sum},
 \begin{lemma*}[Lemma~ \ref{lem:Log_sum}]
For any non-negative real numbers $a_1,\ldots, a_n\in[0,\amax]$, and $\ba \geq 0$,
\begin{align*}
\sum_{i=1}^n \frac{a_i}{\ba+\sum_{j=1}^{i-1} a_j} 
\le 
2+\frac{4\amax}{\ba}+2\log\left( 1+\sum_{i=1}^{n-1} a_i/\ba\right) ~.
\end{align*}
\end{lemma*}
 
 Recalling that $Z_t\in[0,G]$ we may use  the above lemma  to bound term $\rC$,
 \begin{align} \label{eq:TermC_Non}
 \rC
 &:=
 \sum_{t=1}^T \| g_t - M_t\|_* \cdot \|x_t-y_t\|  \nonumber\\
 &\leq
 \sum_{t=1}^T  2G \cdot \sqrt{\|x_t-y_t\|^2 + \|x_t-y_{t-1}\|^2} \nonumber\\
 &=
 2\sqrt{5}G\sum_{t=1}^T \eta_t Z_t \nonumber\\
 &\leq
5 G\sqrt{T} \sqrt{\sum_{t=1}^T \eta_t^2 Z_t^2} \nonumber\\
&\leq
5 GD\sqrt{T} \sqrt{\sum_{t=1}^T  \frac{Z_t^2}{G_0^2+\sum_{\tau=1}^{t-1}Z_\tau^2}} \nonumber\\
 &\leq
 5 GD\sqrt{T} \sqrt{2+4 (G^2/G_0^2)+2\log\left( \frac{G_0^2+\sum_{t=1}^{T-1}Z_t^2}{G_0^2}\right) } \nonumber\\
 &\leq
 5GD \sqrt{T}\sqrt{2+4 (G^2/G_0^2)+ 2\log(1+ TG^2/G_0^2)}~,
 \end{align}
where the second line uses the bound $G$ on the magnitude of the gradients (in the dual norm). The third line uses $Z_t^2: = \left(\|x_t-y_t\|^2 + \|x_t-y_{t-1}\|^2\right)/{5\eta_t^2}$.

\paragraph{Conclusion:} Combining Equations \eqref{eq:TermA_Non}  \eqref{eq:TermB_Non} and
\eqref{eq:TermC_Non}
 into 
Eq.~\eqref{eq:RegGuranteesNonSmooth} implies the following regret bound for Alg.~\ref{alg:UniMP},
$$
\regret \leq 4DG_0+5GD \sqrt{T}\sqrt{2+4 (G^2/G_0^2)+ 2\log(1+ TG^2/G_0^2)}~.
$$
Combining the above with the definition of $\bar{x}_T$ and using Jensen's inequality implies,
$$
\Delta(\bar{x}_T,x)  \leq O\left(\frac{\alpha GD\sqrt{\log T}}{\sqrt{T}}\right)~, \qquad \forall x\in\K~.
$$  
where we used the notation $\alpha: = \max\left\{{G}/{G_0},{G_0}/{G}\right\}$.
 We also used the fact that $\forall x_t,x\in\K$ the following holds,
 $$
 \Delta(x_t,x) \leq F(x)\cdot(x_t-x):= g_t\cdot(x_t-x)~.
 $$ 
 \end{proof}

\newpage
\section{Proofs for the Stochastic Setting (Section~\ref{sec:Stochastic})}
\label{sec:AppStochastic}
\subsection{Proof of Theorem~\ref{thm:NonSmoothNoisy}}
\label{sec:Proof_thm_NonSmoothNoisy}
\begin{proof}
Let us denote by $\tg_t$ the noisy estimates of $g_t: = F(x_t)$, and by $\tilde{M}_t$ the noisy estimates of $M_t:=F(y_{t-1})$.
Recalling the regret bound of  Eq.~\eqref{eq:RegGurantees} implies that for any  $x\in\K$,
\begin{align}\label{eq:RegGuranteesNoisy}
 \sum_{t=1}^T\tg_t&\cdot(x_t-x)  \nonumber\\
&\leq
\frac{D^2}{\eta_1} + \frac{D^2}{\eta_T} + \sum_{t=1}^T \| \tg_t - \tilde{M}_t\|_* \cdot \|x_t-y_t\|
-\frac{1}{2}\sum_{t=1}^T \eta_t^{-1}\left(\|x_t-y_t\|^2 + \|x_t -y_{t-1}\|^2 \right)~.
\end{align}
 Now following the exact steps as in the proof of Theorem~\ref{thm:NonSmooth} implies the following holds w.p.~1,
 $$
 \sum_{t=1}^T\tg_t\cdot(x_t-x)  \leq O(\alpha GD\sqrt{T\log T})~.
 $$
 Recalling the definition of $\bar{x}_T$, and using Jensen's inequality we obtain  for any $x\in\K$,  \begin{align} \label{eq:NoisyDelta}
T\cdot \Delta(\bar{x}_T,x)
& \leq
\sum_{t=1}^T \Delta({x}_t,x)  \nonumber\\
&\leq
 \sum_{t=1}^Tg_t\cdot(x_t-x) \nonumber\\
 & =
  \sum_{t=1}^T\tg_t\cdot(x_t-x) -  \sum_{t=1}^T\zeta_t\cdot(x_t-x) \nonumber\\
   &\leq
O(\alpha  GD\sqrt{T\log T}) -  \sum_{t=1}^T\zeta_t\cdot(x_t-x)~,
 \end{align}
 where we have used the following notation,
 $$
 \zeta_t: = \tg_t-g_t ~.
 $$
 and clearly we have $\E[\zeta_t\vert x_t] =0$, and  $\{ \zeta_t\}_t$ is a martingale difference sequence.
 
 Let $x^*: = \argmax_{x\in\K}\Delta(\bar{x}_t,x)$.  Taking $x=x^*$ and taking expectation over Eq.~\eqref{eq:NoisyDelta} gives,
\begin{align} \label{eq:noisyCaseNonsmooth}
T\cdot \E\Delta(\bar{x}_t,x^*)
& \leq 
O(\alpha GD\sqrt{T\log T}) -  \E\sum_{t=1}^T\zeta_t\cdot(x_t-x^*)  \nonumber\\
&=
O(\alpha GD\sqrt{T\log T})+  \E\sum_{t=1}^T\zeta_t\cdot x^*~.
 \end{align}
 Thus, to establish the proof we are left to show that $\E\sum_{t=1}^T \zeta_t \cdot x^*\leq O(GD\sqrt{T\log T})$. 
We will require the following proposition (its proof appears in Appendix~\ref{sec:Proof_prop:Concentration}),
\begin{proposition} \label{prop:Concentration}
Let $\K \subseteq \reals^d$ be a convex set, and $\R:\K\mapsto \reals$ be a $1$-strongly-convex function w.r.t. a norm $\|\cdot\|$ over $\K$. Also assume that $\forall x\in\K;\;\R(x) - \min_{x\in\K} \R(x)\leq \frac{1}{2}D^2$.
Then for any martingale difference sequence $(Z_i)_{i=1}^n\in \reals^d$, and any random vector $X$ defined over $\K$, we have,
$$ \E \Big[ \Big(  \sum_{i=1}^n Z_i \Big)^\top X  \Big]
\leq
 \frac{D  } { 2  }\sqrt{ \sum_{i=1}^n \E \|Z_i\|_*^2}~,
$$
where $\|\cdot\|_*$ is the dual norm of $\| \cdot\|$.
\end{proposition}
We stress that the theorem  applies for random vectors $X$ which might even dependend of the martingale difference sequence $(Z_i)_{i=1}^n$.

Applying the above lemma with $Z_t  \leftrightarrow \zeta_t$, and $X  \leftrightarrow x^*$ we obtain,
\begin{align*}
\E\sum_{t=1}^T\zeta_t\cdot x^*
&\leq
 \frac{D}{2} \sqrt{ \sum_{t=1}^T \E \|\zeta_t\|_*^2}
 \leq  \frac{D}{2} \sqrt{ 4G^2 T } = DG\sqrt{T}~,
\end{align*}
where we used $\zeta_t = \tg_t -g_t$, together with the bound on the dual norms of $\tg_t,g_t$. 
Combining the above bound inside Eq.~\eqref{eq:noisyCaseNonsmooth} concludes the proof.

 \end{proof}

 \subsection{Proof of Theorem~\ref{thm:SmoothNoisy}}
 \label{Proof_thm:SmoothNoisy}
 Let us denote by $\tg_t$ the noisy estimates of $g_t: = F(x_{t})$, and by $\tilde{M}_t$ the noisy estimates of $M_t: = F(y_{t-1})$.
 
 Recall Equation~\eqref{Eq:rel111} from the proof of Theorem~\ref{thm:Smooth} which states (see Section~\ref{sec:ProofTheoremSmooth}),
  \begin{align*}
  \|g_t-M_t\|_* \cdot \|x_t-y_t\|
  &\leq
   \frac{L }{2}\|x_t-y_{t-1}\|^2 + \frac{L }{2}\|x_t-y_t\|^2~.
 \end{align*}
Using the above together with the triangle inequality we get,
  \begin{align*}
  \|\tg_t-\tM_t\|_* \cdot \|x_t-y_t\|
  &\leq
  \|\xi_t\|_*  \cdot \|x_t-y_t\| + 
   \frac{L }{2}\|x_t-y_{t-1}\|^2 + \frac{L }{2}\|x_t-y_t\|^2~,
 \end{align*}
 where we define,
 $$
 \xi_t: = \tg_t -g_t - (\tM_t -M_t)~.
 $$
 Now, following the exact same analysis as in the proof of Theorem~\ref{thm:Smooth}  (see Section~\ref{sec:ProofTheoremSmooth}) shows the following applies to any $x\in\K$ when using Alg.~\ref{alg:UniMP},
\begin{align}\label{eq:RegretSmoothNoisy}
\sum_{t=1}^T \tg_t\cdot(x_t-x)
&\leq O(\alpha DG+\alpha^2 L  D^2+L  D^2\log(L  D/G_0)) 
+
 \underset{\rD}{\underbrace{\left[\sum_{t=1}^T \|\xi_t\|_*  \cdot \|x_t-y_t\|\right]}}~.
\end{align}
Now similarly to the proof of Theorem~\ref{thm:NonSmoothNoisy} we can show the following to hold for  $x^*: = \argmax_{x\in\K}\Delta(\bar{x}_t,x^*)$ (see Eq.~\eqref{eq:NoisyDelta} and \eqref{eq:noisyCaseNonsmooth}),
\begin{align} \label{eq:noisyCaseSmooth}
T\cdot &\E\Delta(\bar{x}_t,x^*)  \nonumber\\
& \leq 
\E\sum_{t=1}^T \tg_t\cdot(x_t-x) -  \E\sum_{t=1}^T\zeta_t\cdot(x_t-x)  \nonumber\\
& \leq
O(\alpha DG+\alpha^2 L  D^2+L  D^2\log(L  D/G_0))+
\underset{\rD}{\underbrace{\E\left[\sum_{t=1}^T \|\xi_t\|_*  \cdot \|x_t-y_t\|\right]}}
+
\underset{\rE}{\underbrace{\E\sum_{t=1}^T\zeta_t\cdot x^* }}~,
 \end{align}
where we define $\zeta_t: = \tg_g -g_t$.
Next, we will show that both $\rD$ and $\rE$ above are bounded by $O(\sigma D\sqrt{T\log T})$. This will conclude the proof.

\paragraph{ Bounding $\rD$:}
Using Cauchy-Schwarz we have,
\begin{align}\label{eq:VarTerm}
\sum_{t=1}^T \|\xi_t\|_*  \cdot \|x_t-y_t\| \leq \sqrt{\sum_{t=1}^T \|\xi_t\|_*^2}
\sqrt{\sum_{t=1}^T \|x_t-y_t\|^2}~.
\end{align}
Next we show that the sum in the second root is bounded by $O(\log T)$.
Recalling the definition of $Z_t$ we get,
\begin{align}\label{eq:DtermDecomp}
 \sum_{t=1}^T \|x_t-y_t\|^2
& \leq
\sum_{t=1}^T \left(\|x_t-y_t\|^2 +\|x_t-y_{t-1}\|^2 \right) \nonumber \\
&=
5\sum_{t=1}^T \eta_t^2 Z_t^2 \nonumber \\
&=
5D^2 \sum_{t=1}^T \frac{ Z_t^2}{G_0^2+\sum_{\tau=1}^{t-1}Z_\tau^2} \nonumber \\
&\leq
30D^2\max\{1,G^2/G_0^2\} + 10D^2 \log(1+G^2 T/G_0^2)~,
\end{align}
where we used $\eta_t =D/\sqrt{ G_0^2+\sum_{\tau=1}^{t-1}Z_\tau^2 }$, we have also used $\forall t;\;Z_t\in[0,G]$ together with Lemma~\ref{lem:Log_sum}, which we remind below.
\begin{lemma*}[Lemma~\ref{lem:Log_sum}]
For any non-negative real numbers $a_1,\ldots, a_n\in[0,\amax]$, and $\ba \geq 0$,
\begin{align*}
\sum_{i=1}^n \frac{a_i}{\ba+\sum_{j=1}^{i-1} a_j} 
\le 
2+\frac{4\amax}{\ba}+2\log\left( 1+\sum_{i=1}^{n-1} a_i/\ba\right) ~.
\end{align*}
\end{lemma*}
Thus, combining Eq.~\eqref{eq:DtermDecomp} inside Eq.~\eqref{eq:VarTerm}, and taking expectation we conclude that,
\begin{align} \label{eq:DboundSmooth}
\rD:
& = 
\E\left[\sum_{t=1}^T\|\xi_t\|_*  \cdot \|x_t-y_t\|\right] 
 \leq 6D\max\{1,G_0/G_0\}\sqrt{1+\log T}\cdot  \E\sqrt{\sum_{t=1}^T \|\xi_t\|_*^2} \nonumber\\
&\leq
12\alpha D\sigma \sqrt{T(1+\log T)}~.
\end{align}
where we have used Jensen's inequality with respect to the $H(u) : =\sqrt{u}$, as well as $\E\|\xi_t\|_*^2\leq 4\sigma^2$. We also used $\alpha: = \max\{G/G_0,G_0/G \}$.

\paragraph{ Bounding $\rE$:}
Using Proposition~\ref{prop:Concentration} (see section~\ref{sec:Proof_thm_NonSmoothNoisy}), and taking 
$Z_t  \leftrightarrow \zeta_t$, and $X  \leftrightarrow x^*$ we obtain,
\begin{align}\label{eq:smoothVariaceBound}
\E\sum_{t=1}^T\zeta_t\cdot x^*
&\leq
 \frac{D}{2} \sqrt{ \sum_{t=1}^T \E \|\zeta_t\|_*^2}
 \leq  \frac{D}{2} \sqrt{ \sigma^2 T } = D\sigma\sqrt{T}~,
\end{align}
where we used $\zeta_t = \tg_t -g_t$, together with the bound on the variance of $\tg_t$.

\paragraph{ Concluding:} 
Using the bounds in Eq.~\eqref{eq:DboundSmooth} and \eqref{eq:smoothVariaceBound}, inside Eq.~\eqref{eq:noisyCaseSmooth} gives,
\begin{align}
T\cdot &\E\Delta(\bar{x}_t,x^*)  \nonumber\\
& \leq
O(\alpha DG+\alpha^2 L  D^2+L  D^2\log(L  D/G_0))+ O(\alpha\sigma D\sqrt{T\log T})~,
 \end{align}
which concludes the proof.

\subsection{Proof of Theorem~\ref{prop:Concentration}}
\label{sec:Proof_prop:Concentration}
\begin{proof}
Let us denote by $\R^*$ the Fenchel dual of $\R$.
Thus the $1$-strong-convexity of $\R$ w.r.t. $\|\cdot\|$ implies that $\R^*$ is $1$-smooth w.r.t. $\|\cdot\|_*$.

We also denote by $x_0 \in \K$ be the minimizer of $\R$ on $\K$, which we assume to be in the relative interior of $\K$, so that $\nabla \R(x_0)=0$. Without loss of generality, we also assume that $\R(x_0)=0$. This implies that $\R^\ast(0) := \max_{x\in\reals^d}\{x\cdot 0 - \R(x)\} = 0 $.

Now recall the  Fenchel-Young inequality,
\begin{lemma}[Fenchel-Young inequality]
Let $f:\K\mapsto \reals$ be a convex function, and $f^*:\reals^d\mapsto \reals$ be its Fenchel dual, then
$$
f(x) + f^*(y) \geq x^\top y,\qquad \forall x\in\text{dom}(f), y\in\text{dom}(f^*)~.
$$
\end{lemma}
Using this inequality and taking  $ y = s \sum_{i=1}^n Z_i$ and $x=X $, we have,
\begin{align}\label{eq:FN_INEQ}
\Big(  \sum_{i=1}^n Z_i \Big)^\top X 
= \frac{1}{s} \Big( s \sum_{i=1}^n Z_i \Big)^\top X
\leq \frac{1}{s} \Big(  \underset{\rA}{\underbrace{\R(X)}} + \R^\ast\big(  \underset{\rB}{\underbrace{s \sum_{i=1}^n Z_i\big)}} \Big)
~.
\end{align}
Next we bound the two terms in the above inequality.\\
\textbf{Bounding $\rA$:}
Using $\R(x_0) = 0$ together with the boundedness of $\R$, gives,
$$
\rA: = \R(X) =  \R(X)- \R(x_0) \leq \frac{1}{2}D^2~.
$$
\textbf{Bounding $\rB$:} To bound this term we first require the following lemma that  can be found in \cite{juditsky2008large,Kakade13} (for completeness we provide a proof in Appendix~\ref{app:Proof_lem:KakadeConcentration}),
\begin{lemma} \label{lem:KakadeConcentration}
Let $f:\reals^d$ be a $L $-smooth function w.r.t. a norm $\|\cdot\|_*$, such that $f(0) = 0$. Also let $(M_i)_{i}$ be a martingale difference sequence. Then the following holds,
$$
\E f\big(  \sum_{i=1}^n M_i\big)
\leq   \frac{L }{2}\sum_{i=1}^n \E \|M_i\|_*^2~.
$$
\end{lemma}
Applying the above lemma with $f \leftrightarrow  \R^*$, and $M_i \leftrightarrow  sZ_i$ gives,
$$
\E \R^\ast\big( s \sum_{i=1}^n Z_i\big)
\leq  {  \frac{s^2  }{2 }} \sum_{i=1}^n \E \|Z_i\|_*^2~.
$$

Combining the bounds on $\rA$ and $\rB$ inside Eq.~\eqref{eq:FN_INEQ} and taking expectation gives,
$$ \E \Big[ \Big(  \sum_{i=1}^n Z_i \Big)^\top X  \Big]
\leq
\frac{1}{s} ( \frac{1}{2} D^2 {  + \frac{s^2  }{2 }\sum_{i=1}^n \E \|Z_i\|_*^2} )
= \frac{D^2}{2 s} + {  \frac{s  }{2 }} \sum_{i=1}^n \E \|Z_i\|_*^2~.
$$
Taking $s = \frac{D   }{\sqrt{ \sum_{i=1}^n \E \|Z_i\|_*^2}}$, we get,
$$ \E \Big[ \Big(  \sum_{i=1}^n Z_i \Big)^\top X  \Big]
\leq
 \frac{D  } { 2  }\sqrt{\sum_{i=1}^n \E \|Z_i\|_*^2}~,
$$
which concludes the proof.
\end{proof}

\subsubsection{Proof of Lemma~\ref{lem:KakadeConcentration}}
\label{app:Proof_lem:KakadeConcentration}
\begin{proof}
We will prove the lemma by induction over $n$.
For the base case $n=1$, we may use the smoothness of $f$ to get,
$$
f(Z_1) = f(Z_1) -f(0) \leq \nabla f(0)^\top Z_1 + \frac{L }{2}\|Z_1\|_*^2~.
$$
Taking expectation and using $\E Z_1 = 0$ the lemma follows.

Now for the induction step, assume that $\E f(\sum_{i=1}^{n-1} Z_i)\leq \frac{L }{2}\sum_{i=1}^{n-1}\E\| Z_i\|_*^2$.  Using the smoothness of $f$  gives,
\begin{align*}
f(\sum_{i=1}^n Z_i)& = f(Z_n + \sum_{i=1}^{n-1} Z_i)\\
& \leq
  f( \sum_{i=1}^{n-1} Z_i) +\nabla f\left(\sum_{i=1}^{n-1} Z_i\right)^\top Z_n
+\frac{L }{2} \|Z_n\|_*^2~.
\end{align*}
Taking expectation and using $\E[ Z_n\vert Z_1,\ldots, Z_{n-1}]= 0$, as well as the induction assumption establishes the lemma.
\end{proof}

 \newpage

\section{Additional Proofs}
\label{sec:AppC}

\subsection{Proof of Lemma~\ref{lem:RakhlinSridharan}}
 \label{sec:Proof_lem:RakhlinSridharan}
 \begin{proof}
 For any $x^*\in\K$,
 \begin{align}\label{eq:Optimistic1}
 g_t \cdot (x_t -x^*)
 & =g_t\cdot(x_t-y_t) +  g_t\cdot(y_t-x^*)  \nonumber\\
 &=(g_t-M_t)\cdot(x_t-y_t) + M_t \cdot(x_t-y_t) +  g_t\cdot(y_t-x^*) ~.
 \end{align}
 Moreover, by Cauchy-Schwarz,
 \begin{align}\label{eq:Optimistic2}
(g_t-M_t)\cdot(x_t-y_t) 
\leq 
\|g_t-M_t\|_* \|x_t-y_t\|~.
 \end{align}
Also, any update of the form $a^* = \argmin_{a\in A} a\cdot z + \Dr(a,c)$, satisfies for any $b\in A$,
\begin{align*}
z\cdot (a^* -b)  \leq \Dr(b,c) - \Dr(b,a^*)- \Dr(a^*,c)~.
 \end{align*}
 Combining this with the Optimistic OGD learning rule (Eq.~\eqref{eq:OptimisticGD}) gives, 
 (taking $a^* \leftrightarrow x_t,\; b \leftrightarrow y_t,\;c \leftrightarrow y_{t-1},\; z\leftrightarrow M_t$),
 \begin{align}
 M_t \cdot(x_t-y_t) \leq  \frac{1}{\eta_t} \left( \Dr(y_t,y_{t-1}) - \Dr(y_t,x_t)- \Dr(x_t,y_{t-1}) \right)~,
 \end{align}\label{eq:Optimistic3}
 as well as
(taking $a^* \leftrightarrow y_t,\; b \leftrightarrow x^*,\;c \leftrightarrow y_{t-1},\; z\leftrightarrow g_t$),
 \begin{align}\label{eq:Optimistic4}
 g_t \cdot(y_t-x^*) \leq   \frac{1}{\eta_t} \left(  \Dr(x^*,y_{t-1}) - \Dr(x^*,y_t)- \Dr(y_t,y_{t-1}) \right)~.
 \end{align}
 Combining Equations~\eqref{eq:Optimistic2}-\eqref{eq:Optimistic4} inside Eq.~\eqref{eq:Optimistic1}
we obtain, 
\begin{align*}
 g_t &\cdot (x_t -x^*) \\
 &\leq
\|g_t-M_t\|_* \|x_t-y_t\| + \frac{1}{\eta_t}\left(\Dr(x^*,y_{t-1})- \Dr(x^*,y_t) - \Dr(y_t,x_t)- \Dr(x_t,y_{t-1})\right) \\
&\leq
\|g_t-M_t\|_* \|x_t-y_t\| + \frac{1}{\eta_t}\left(\Dr(x^*,y_{t-1})- \Dr(x^*,y_t) - \frac{1}{2}\|x_t-y_t\|^2- \frac{1}{2}\|x_t-y_{t-1}\|^2\right)~,
 \end{align*}
 where the last line uses the $1$-strong-convexity of $\R$, implying that $\forall x,y\in\K;\;\Dr(x,y)\geq \frac{1}{2}\|x-y\|^2$.  Summing over $t\in[T]$ we obtain that for any $x^*\in\K$,
 \begin{align*}
 \sum_{t=1}^T& g_t\cdot (x_t-x^*) \\
 &\leq
   \frac{1}{\eta_1}\Dr(x^*,y_{0})
 +\sum_{t=2}^T\Dr(x^*,y_{t-1})\left(\frac{1}{\eta_{t}} -\frac{1}{\eta_{t-1}} \right) \\
 &\quad
 + \sum_{t=1}^T \| g_t - M_t\|_* \cdot \|x_t-y_t\|
-\frac{1}{2}\sum_{t=1}^T \eta_t^{-1}\left(\|x_t-y_t\|^2 + \|x_t -y_{t-1}\|^2 \right) \\
&\leq
   \frac{D^2}{\eta_1} 
 + D^2 \sum_{t=2}^T\left(\frac{1}{\eta_{t}} -\frac{1}{\eta_{t-1}} \right) \\
 &\quad
 + \sum_{t=1}^T \| g_t - M_t\|_* \cdot \|x_t-y_t\|
-\frac{1}{2}\sum_{t=1}^T \eta_t^{-1}\left(\|x_t-y_t\|^2 + \|x_t -y_{t-1}\|^2 \right) \\
&\leq
\frac{D^2}{\eta_1} +\frac{D^2}{\eta_T}  \\
 &\quad
 + \sum_{t=1}^T \| g_t - M_t\|_* \cdot \|x_t-y_t\|
-\frac{1}{2}\sum_{t=1}^T \eta_t^{-1}\left(\|x_t-y_t\|^2 + \|x_t -y_{t-1}\|^2 \right)~,
 \end{align*}
where in the second inequality we use the fact that $\eta_t$ is monotonically non-increasing and thus
$\frac{1}{\eta_{t}} - \frac{1}{\eta_{t-1}}\geq 0$, as well as $\forall x\in\K;\;\Dr(x,y_{0})\in [0,D^2]$. This concludes the proof.
 
 \end{proof}
\subsection{Proof of Lemma~\ref{lem:SqrtSumReversedGeneralized}}
\label{sec:Proof_SqrtSumReversedGeneralized}
\begin{proof} 
\textbf{First direction:}
Here we actually prove a stronger result which is the following,
\begin{align}\label{eq:LowerSumStronger}
\sqrt{\ba+\sum_{i=1}^{n} a_i} -\sqrt
{\ba}\leq \sum_{i=1}^n \frac{a_{i}}{\sqrt{\ba+\sum_{j=1}^{i-1} a_j}}~.
\end{align}
The above combined with $a_n\geq0$ immediately implies the first part of 
 Lemma~\ref{lem:SqrtSumReversedGeneralized}.

We will prove this Eq.~\eqref{eq:LowerSumStronger}  by induction.
The base case, $n=1$,  holds since in this case,
$$
\sum_{i=1}^1 \frac{a_{i}}{\sqrt{\ba+\sum_{j=1}^{i-1} a_j}} = \frac{a_1}{\sqrt{\ba}} \geq \sqrt{\ba+a_1}-\sqrt{\ba}~,
$$
The above is equivalent to,
$$
(\ba+a_1)^2 \geq \ba(\ba+a_1)~,
$$
which is holds true since $a_1,\ba\geq0$.

For the induction step assume that the Eq.~\eqref{eq:LowerSumStronger} holds for $n-1$ and let us show it holds for $n$.
By the induction assumption,
\begin{align*}
\sum_{i=1}^n \frac{a_{i}}{\sqrt{\ba+\sum_{j=1}^{i-1} a_j}}
&\geq
\sqrt{\ba+\sum_{i=1}^{n-1}a_i} -\sqrt{\ba}+ \frac{a_n}{\sqrt{\ba+\sum_{i=1}^{n-1} a_i}}\\
& = \sqrt{Z-x} + \frac{x}{\sqrt{Z-x}}-\sqrt{\ba}~,
\end{align*}
where we denote $x:= a_n$ and $Z = \ba+\sum_{i=1}^n a_i$ (note that $x< Z$).
Thus, in order to prove the lemma it is sufficient to show that,
$$
\sqrt{Z-x} + \frac{x}{\sqrt{Z-x}} \geq \sqrt{Z}~.
$$
Looking at the function $H(x) := \sqrt{Z-x} + \frac{x}{\sqrt{Z-x}}$ it is immediate to validate that $H(\cdot)$ is monotonically increasing for any $x\in[0,Z]$ (since its derivative is non-negative in this line segment) and therefore for any $x\in[0,Z]$ we have,
$$
\sqrt{Z-x} + \frac{x}{\sqrt{Z-x}} = H(x)\geq H(0) = \sqrt{Z}~.
$$
This establishes Eq.~\eqref{eq:LowerSumStronger} which in turn concludes  the first part of the proof.

\paragraph{Second direction:} 
For this part of the proof we will need the following lemma which we prove in Section~\ref{app:Prooflem:SqrtSumReversed},
\begin{lemma}\label{lem:SqrtSumReversed}
For any non-negative numbers $a_1,\ldots, a_n\in [0,\amax]$, the following holds:
\begin{equation*}
\sum_{i=1}^n \frac{a_{i}}{\sqrt{\amax+\sum_{j=1}^{i-1} a_j}} \leq 
 2\sqrt{\amax} +2\sqrt{\amax +\sum_{i=1}^{n-1} a_i}~. 
\end{equation*}
\end{lemma}
Now let us divide into two cases. 
\textbf{Assume $a\leq \ba$},
in this case it is clear that,
\begin{equation*}
\sum_{i=1}^n \frac{a_{i}}{\sqrt{\ba+\sum_{j=1}^{i-1} a_j}} 
\leq
\sum_{i=1}^n \frac{a_{i}}{\sqrt{\amax+\sum_{j=1}^{i-1} a_j}} 
\leq 
 2\sqrt{\amax} +2\sqrt{\amax +\sum_{i=1}^{n-1} a_i} 
 \leq 
 2\sqrt{a} +2\sqrt{\ba +\sum_{i=1}^{n-1} a_i}~,
\end{equation*}
where the second inequality holds by Lemma~\ref{lem:SqrtSumReversed}.
We are therefore left to analyze the case \textbf{where  $\ba \leq a$}. In this case, let us denote the following,
$$
N_0 = \min \left\{i\in[n]: \sum_{j=1}^{i-1}a_j \geq \amax  \right\}~.
$$
Next we divide the relevant sum according to $N_0$,
\begin{align*}
 \sum_{i=1}^n \frac{a_{i}}{\sqrt{\ba+\sum_{j=1}^{i-1} a_j}} &=
 \sum_{i=1}^{N_0-1} \frac{a_{i}}{\sqrt{\ba+\sum_{j=1}^{i-1} a_j}} 
 +
 \sum_{i=N_0}^{n} \frac{a_{i}}{\sqrt{\ba+\sum_{j=1}^{i-1} a_j}} \\
 &\leq
 \frac{1}{\sqrt{\ba}} \sum_{i=1}^{N_0-1}a_{i} 
  +
\sum_{i=N_0}^{n} \frac{a_{i}}{\sqrt{\sum_{j=1}^{i-1} a_j}} \\
 &\leq
\frac{ 2\amax}{\sqrt{\ba}}
 + 
 \sum_{i=N_0}^{n} \frac{a_{i}}{\sqrt{\frac{1}{2}\amax + \frac{1}{2}\sum_{j=1}^{i-1} a_j}} \\
 &\leq
\frac{ 2\amax}{\sqrt{\ba}}
 + 
\sqrt{2} \sum_{i=N_0}^{n} \frac{a_{i}}{\sqrt{\amax + \sum_{j=N_0}^{i-1} a_j}} \\
&\leq
\frac{ 2\amax}{\sqrt{\ba}}
 + 
 2\cdot\sqrt{2}\sqrt{\amax} +  2\cdot\sqrt{2}\sqrt{\amax + \sum_{j=N_0}^{n-1} a_j}\\
&\leq
\frac{ 2\amax}{\sqrt{\ba}}
 +
 3\sqrt{\amax} +  3\sqrt{\ba + \sum_{j=1}^{n-1} a_j}~,
\end{align*}
where we used $\sum_{j=1}^{N_0-2} a_j \leq \amax$, as well as  $\forall i\geq N_0;~\sum_{j=1}^{i-1}a_j \geq \amax$. Both follow by the definition of $N_0$. The last line uses  $\amax\leq \sum_{j=1}^{N_0-1}a_j$, and the line before last (i.e., the fifth line) uses Lemma~\ref{lem:SqrtSumReversed}.
This concludes  the second part of the proof.
\end{proof}

\subsection{Proof of Lemma~\ref{lem:SqrtSumReversed}}
\label{app:Prooflem:SqrtSumReversed}
\begin{proof}
We will require the following lemma from  \cite{mcmahan2010adaptive} (appear as Lemma $7$ therein).
 \begin{lemma}\label{lem:SqrtSumOriginal}
For any non-negative numbers $a_1,\ldots, a_n$ the following holds:
\begin{equation*}
\sum_{i=1}^n \frac{a_i}{\sqrt{\sum_{j=1}^i a_j}} \leq 2\sqrt{\sum_{i=1}^n a_i}~.
\end{equation*}
\end{lemma}

Using the above lemma together with $\forall i;~a_i\leq \amax$ we have,
\begin{align*}
\sum_{i=1}^n \frac{a_{i}}{\sqrt{\amax+\sum_{j=1}^{i-1} a_j}} 
&\leq 
\sum_{i=1}^n \frac{a_{i}}{\sqrt{\sum_{j=1}^{i} a_j}} \\
&\leq
2\sqrt{\sum_{i=1}^n a_i} \\
&\leq
2\sqrt{\sum_{i=1}^{n-1} a_i} + 2\sqrt{\amax} \\
&\leq
2\sqrt{\amax +\sum_{i=1}^{n-1} a_i} + 2\sqrt{\amax}~,
\end{align*}
where we used $\sqrt{a+b}\leq \sqrt{a}+\sqrt{b}$ which hold $\forall a,b\geq 0$.
This concludes the second part of the proof.
\end{proof}

\subsection{Proof of Lemma~\ref{lem:Log_sum}}
\label{app:Prooflem:Log_sum}
\begin{proof}
Let us denote the following:
$$
N_0 = \min \left\{i\in[n]: \sum_{j=1}^{i-1}a_j \geq \amax  \right\}~.
$$
Next we divide the relevant sum according to $N_0$,
\begin{align}\label{eq:Sum1Logsum}
\sum_{i=1}^n \frac{a_i}{\ba+\sum_{j=1}^{i-1} a_j} 
&=
\sum_{i=1}^{N_0-1} \frac{a_i}{\ba+\sum_{j=1}^{i-1} a_j} 
+
\sum_{i=N_0}^{n} \frac{a_i}{\ba+\sum_{j=1}^{i-1} a_j}   \nonumber\\
&\leq
\frac{1}{\ba}\sum_{i=1}^{N_0-1} a_i
+
\sum_{i=N_0}^{n} \frac{a_i}{\frac{1}{2}\ba+\frac{1}{2}\amax + \frac{1}{2}\sum_{j=1}^{i-1} a_j}  \nonumber\\
&\leq
\frac{2\amax}{\ba}
 + 
2\sum_{i=N_0}^{n} \frac{a_i}{\ba +a_i+ \sum_{j=N_0}^{i-1} a_j}  \nonumber\\
&=
\frac{2\amax}{\ba}
 + 
2\sum_{i=N_0}^{n} \frac{a_i/\ba}{1 + \sum_{j=N_0}^{i} a_j/\ba}  \nonumber\\
&\leq
\frac{2\amax}{\ba} + 2 + 2\log\left( 1+\sum_{i=N_0}^{n} a_i/\ba\right)  \nonumber\\
&\leq
\frac{2\amax}{\ba} + 2 + 2\log\left( 1+\sum_{i=1}^{n} a_i/\ba\right)~,
\end{align}
where we used $\sum_{j=1}^{N_0-2} a_j \leq \amax$, as well as  $\forall i\geq N_0;~\sum_{j=1}^{i-1}a_j \geq \amax$. Both follow by the definition of $N_0$. And the fourth line uses the following lemma which we borrow  from \cite{levy2018online} (appears as Lemma $\textbf{A.3}$ therein),
\begin{lemma} \label{lem:Log_sumOriginal}
For any non-negative real numbers $b_1,\ldots, b_n$,
\begin{align*}
\sum_{i=1}^n \frac{b_i}{1+\sum_{j=1}^i b_j} 
\le 
1+\log\left( 1+\sum_{i=1}^n b_i\right)~.
\end{align*}
\end{lemma}
Now notice that,
\begin{align}
\label{eq:Logsum2}
\log\left( 1+\sum_{i=1}^{n} a_i/\ba\right)
&=
 \log\left(1+\sum_{i=1}^{n-1} a_i/\ba\right) + \log\left( 1+\frac{a_n/\ba}{1+\sum_{i=1}^{n-1} a_i/\ba}\right)\nonumber\\
&\leq
\log\left( 1+\sum_{i=1}^{n-1} a_i/\ba\right)  + \frac{a_n/\ba}{1+\sum_{i=1}^{n-1} a_i/\ba} \nonumber\\
&\leq
\amax/\ba+  \log\left( 1+\sum_{i=1}^{n-1} a_i/a\right)~,
\end{align}
where the third line uses $\forall x\geq0,~\log(1+x)\leq x$, and the last line uses $a_n\leq \amax$.
Combining Equations~\eqref{eq:Sum1Logsum} and~\eqref{eq:Logsum2} concludes the proof. 
\end{proof}

\end{document}